\documentclass[sigconf]{acmart}
\renewcommand\footnotetextcopyrightpermission[1]{}

\usepackage{graphicx}
\usepackage{amsmath}
\usepackage{makecell}
\usepackage{multirow}
\usepackage{wrapfig}
\usepackage{pifont} 
\usepackage{colortbl}
\usepackage{algorithm}
\usepackage{algpseudocode}
\usepackage{listings}
\usepackage{color}
\definecolor{mygray}{RGB}{230, 230, 230}
\definecolor{mygreen}{rgb}{0.92, 1.0, 0.92}
\usepackage{array}
\usepackage{graphicx} 

\usepackage{booktabs}   % \toprule, \midrule, \bottomrule
\usepackage{xcolor}     % colors like cyan!10
\usepackage{array}      % p{width} column type and nicer column tools
\usepackage[T1]{fontenc} % better font encoding (recommended)

\AtBeginDocument{%
  }

\setcopyright{acmlicensed}
\copyrightyear{2026}
\acmYear{2026}
\acmDOI{XXXXXXX.XXXXXXX}
\begin{document}

\title{Text-Augmented Multimodal LLMs for Chemical Reaction Condition Prediction}


\newcommand*{\affaddr}[1]{#1} % No op here. Customize it for different styles.
\newcommand*{\affmark}[1][\dagger]{\textsuperscript{#1}}
\author{
\textbf{Yu Zhang}\affmark[1]{$^*$}, \ \textbf{Ruijie Yu}\affmark[1]{$^*$}, \ \textbf{Kaipeng Zeng}\affmark[1], \  \textbf{Ding Li}\affmark[1], \ \textbf{Feng Zhu}\affmark[2], \ \\ \textbf{Xiaokang Yang}\affmark[1], \ \textbf{Yaohui Jin}\affmark[1]{$^\dagger$}, \ \textbf{Yanyan Xu}\affmark[1]{$^\dagger$}\\
\affaddr{\affmark[1]MoE Key Laboratory of Artificial Intelligence, AI Institute, Shanghai Jiao Tong University \\
\affmark[2]Frontiers Science Center for Transformative Molecules (FSCTM), Shanghai Jiao Tong University \\
{$^*$} Equal contribution; \ \ {$^\dagger$}\  Corresponding authors, \{jinyh, yanyanxu\}@sjtu.edu.cn 
\ \ \ }
}

\renewcommand{\shortauthors}{Zhang et al.}

\begin{abstract}
  % A clear and well-documented \LaTeX\ document is presented as an
  % article formatted for publication by ACM in a conference proceedings
  % or journal publication. Based on the ``acmart'' document class, this
  % article presents and explains many of the common variations, as well
  % as many of the formatting elements an author may use in the
  % preparation of the documentation of their work.

  Identifying reaction conditions that are broadly applicable across diverse substrates is a longstanding challenge in chemical and pharmaceutical research. While many methods are available to generate conditions with acceptable performance, a universal approach for reliably discovering effective conditions during reaction exploration is rare. Consequently, current reaction optimization processes are often labor-intensive, time-consuming, and costly, relying heavily on trial-and-error experimentation. Nowadays, large language models (LLMs) are capable of tackling chemistry-related problems, such as molecule design and chemical reasoning tasks. Here, we report the design, implementation and application of \textbf{Chemma-RC}, a text-augmented multimodal LLM to identify effective conditions through task-specific dialogue and condition generation. Chemma-RC learns a unified representation of chemical reactions by aligning multiple modalities—including text corpus, reaction SMILES, and reaction graphs—within a shared embedding module. Performance benchmarking on datasets showed high precision in identifying optimal conditions, with up to 17\% improvement over the current state-of-the-art methods. A palladium-catalysed imidazole C–H arylation reaction was investigated experimentally to evaluate the functionalities of the Chemma-RC in practice. Our findings suggest that Chemma-RC holds significant potential to accelerate high-throughput condition screening in chemical synthesis.
\end{abstract}




\maketitle

\section{Introduction} \label{sec:intro}
% \vspace{-3pt}
% chermistry background
%LLM is a potential solution for AI Chem
% Why LLM can work , introduce our work, Chemma-RC
%

% chemistry background
Chemical synthesis is a crucial step for the discovery of transformative molecules in multiple fields, including drug design, materials, renewable energy, etc.
%high-throughput reaction condition (RC) screening, an effective workflow, exerts an important function in synthesizing molecules~\cite{macarron2011impact}.
%合成方法学表现在只能在很少的底物work，且需要很多实验；高通量、自动化的发展会缓解传统合成方法学的反复试错
%工具辅助的合成设计方法有用，但不通用,容易陷入局部最优。
% 
In chemical synthesis, reaction conditions are usually optimized to maximize the yield of each target molecule and to be applicable to a wide variety of substrates~\cite{shields2021bayesian,taylor2023brief}. Despite synthetic methods have achieved significant advancements over the past few decades, discovering effective reaction conditions from the vast substrates still relies on the trial-and-error experimental efforts~\cite{angello2022closed}. While automated platforms have increased the efficiency of reaction optimization and reduced exploration costs, challenges in exploring reaction conditions still hinder the adoption of new methodologies in synthetic chemistry. Additionally, optimization is often necessary for different target substrates, and pharmaceutically relevant molecules with high structural complexity may not be compatible with existing conditions. Recently, chemists have focused on designing reliable computer-aided synthesis planning (CASP) tools to facilitate condition screening~\cite{corey1969computer, mikulak2020computational, schwaller2021mapping}. Transformer-based models built upon SMILES or reaction graphs have demonstrated strong effectiveness in addressing various chemical tasks~\cite{gao2018using, schwaller2019molecular, andronov2023reagent}. However, practical prediction for reaction conditions is more complex than what can be achieved by using graph methods alone. Therefore, a universal approach to discover effective conditions is still rare~\cite{mehr2020universal,rohrbach2022digitization}. \textit{In summary, to realize efficient synthesis in chemistry, there is an urgent need to facilitate high-efficiency reaction condition prediction.}

% 
% practical answers for molecular design are more complex than what can be a

% The generation begins with a paragraph describing the intended molecule for multi-conditional generation, followed by retrosynthetic planning, detailing each synthesis step—one reaction per paragraph—in reverse order, from the target molecule to purchasable reactants. Thus, multimodal LLMs (MLLMs) are essential, with LLMs handling text generation and graph models managing molecular design.

% Therefore, in this work, we leverage multiple data modalities in chemistry to learn a unified representation for reaction condition prediction.
% The capability to encode different reactions is critical for prediction, as even minor variations in a substrate's functional group can result in fundamentally different reaction conditions. 

\begin{figure*}[!t]
  \centering
  \includegraphics[width=0.9\linewidth]{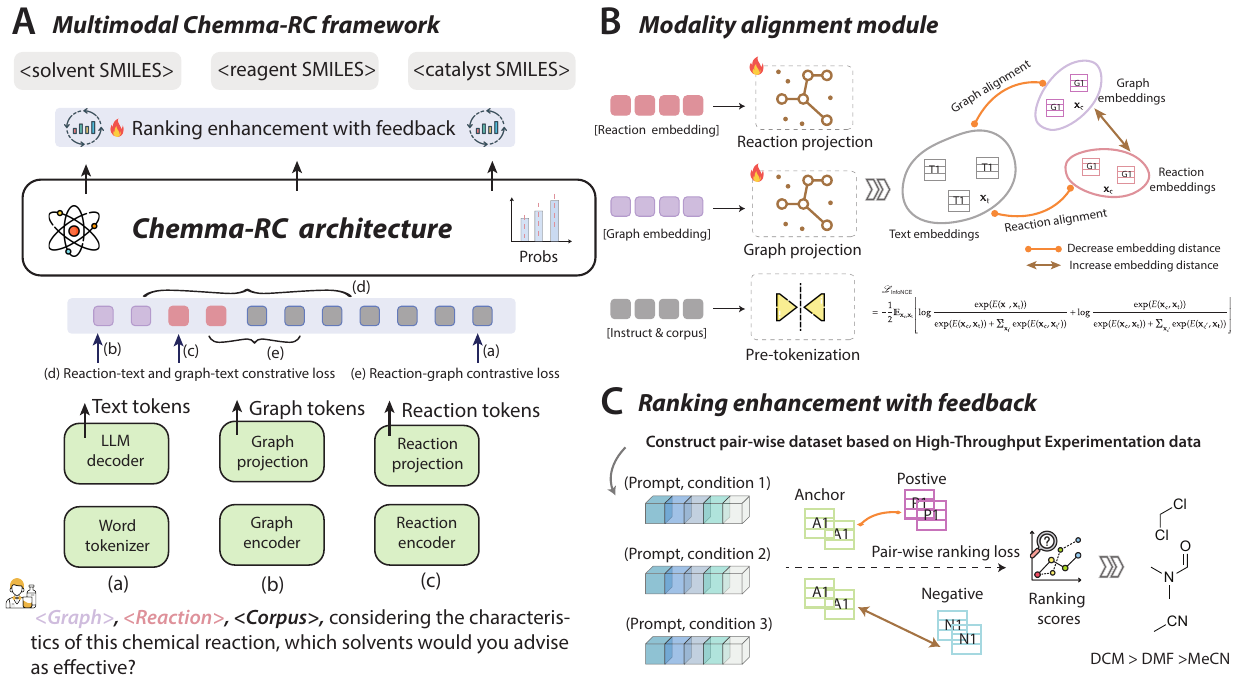}
  \vspace{-10pt}
  \caption{\textbf{Architecture of Chemma-RC.} It processes task-specific questions and generates answers via a two-stage training framework: multimodal supervised fine-tuning followed by post-fine-tuning.
  }
  \label{fig:framework}
% \vspace{-20pt}
\end{figure*}

% lmm
% 现有LLM无法解决表征增强和寻找最优的反应，所以研究化学多模态大模型。化学领域的多模态有什么挑战：1. 在化学领域有一对多的问题，导致通用LMM无法解决；2. 化学数据跨尺度，导致现有对齐方法无法很好实现。针对这两个挑战我们xx。
Nowadays, the emergence of generative large language models (LLMs) or large multimodal models (LMMs), typified by GPT-4 and GPT-4o~\cite{achiam2023gpt}, has sparked significant interest in the field of AI for chemistry~\cite{baum2021artificial, openai2023gpt, boiko2023autonomous, guo2023can, m2024augmenting}. The prediction and design of reaction conditions necessitate LLMs to be controllable for generating molecular structures that satisfy the substrates and synthesizability requirements. These requirements can be articulated as questions for LLM input, as illustrated in Figure~\ref{fig:framework}A. Answering these questions demands a comprehensive understanding of chemical reactions and the relationship between substrates and conditions. However, sequence-based LLMs struggle with this because they are pre-trained or fine-tuned solely on texts. Notably, even in comparatively easier tasks related to molecules, such as captioning and understanding, the best LLMs perform worse than the domain-specific model, like GraphQA, an effective graph-based method in the design of molecules~\cite{gao2022sample}. As we investigate that there are various types of data in the field of chemistry, including simplified molecular-input line-entry system (SMILES)~\cite{weininger1989smiles}, reaction graphs, and a textual corpus of reaction~\cite{schlichtkrull2018modeling}, which encompasses the descriptions of reaction processes and reaction mechanisms. Among several data modalities, chemical large multimodal models (LMMs) are essential, with LLMs handling text generation and domain-specific models managing reaction representations~\cite{livne2024nach0}. However, under the paradigm of LMMs, there are still two important challenges in chemical reaction prediction. First, the inherently `one-to-many' nature of chemical reactions, where a single substrate may correspond to multiple valid reaction conditions, makes it difficult for LMMs to identify optimal reaction conditions. Second, multiple scales of different modalities of data, from atom-level structures to high-level corpora texts~\cite{qian2023predictive}, render conventional cross-modal alignment methods ineffective. Addressing these challenges is essential for building chemical LMMs capable of advancing reaction prediction and optimization. Thus, we propose the multimodal LLM, Chemma-RC, for reaction condition prediction. As shown in Figure~\ref{fig:framework}, Chemma-RC integrates LLMs and the other chemical domain-specific models within a multi-modal auto-regressive framework. It predicts the next token across both word and chemical space, enabling the direct generation of reaction conditions. In summary, we think Chemma-RC can be a potential solution due to the following advantages: (i) foundational LLMs can learn relationships between molecules and reactions, thereby acquiring chemical knowledge akin to the learning process of chemists; (ii) via learning the joint representation of chemical reactions from different modalities--graphs, reaction SMILES, and corpus, LLMs might be better equipped to capture the underlying chemical semantics of reactions, thereby improving the accuracy of reaction condition prediction.

% 最后一段，contribution：（1）通过这些手段我们构造了首个化学多模态大模型，解决多模态表征增强，从而能够找到最优反应体系。（2）两个hightlight 讲LMM应用在化学。（3）实验效果。额外的分析证明模态融合方法的效果。
% Chemma-RC jointly learns a unified reaction representation from SMILES, graphs, and textual corpus for condition prediction. 
The contributions of this work are summarized as follows: 
\begin{enumerate}
    \item we first design a multimodal LLM, Chemma-RC, to jointly learn a unified reaction representation from SMILES, graphs, and corpus for condition prediction;
    \item We design a post-fine-tuning strategy that integrates a ranking enhancement with feedback module to facilitate the generation of optimal conditions;
    \item We design a cross-modality contrastive learning strategy to achieve unified and semantically representations across different modalities.
\end{enumerate}
Through evaluation on benchmark datasets, Chemma-RC exhibits strong generalization capabilities on out-of-domain (OOD) and high-throughput experimentation (HTE) datasets.

\section{Related Work} \label{sec:relation}
% High-throughput reaction condition screening is important but hard,
% (CASP) how traditional methods ，but little focus on rcr
%LLMs
% LLMs for molecule science, and multimodal LLMs for chemical reaction.

% tradition work for RCR
Since the emergence of DeepSeek~\cite{liu2024deepseek}, GPT-4 series~\cite{achiam2023gpt}, LLMs have become foundational models in addressing text-based challenges. The influence of these models is increasingly evident in the fields of chemistry~\cite{huang2024chemeval}, biology, and materials science, where they are being applied to complex molecular studies~\cite{fang2023mol}. In the field of chemistry, Livne et al. introduced a new foundation model, nach0, to solve various chemical and biological tasks, including biomedical question answering, molecular generation~\cite {livne2024nach0}. Zhao et al. propose ChemDFM-X, a large multimodal model that serves as a generalist model to understand five modalities in chemistry~\cite{zhao2024chemdfm, zhao2025developing}. Further, in the study of organic synthesis, reaction conditions are usually designed and optimized to maximize the yield of each target molecule or minimize the cost of the reaction process~\cite{shields2021bayesian,taylor2023brief}. High-throughput condition screening, as an important tool in synthesizing molecules, exerts an important influence on chemical synthesis. 
% However, discovering suitable reaction conditions from the extensive matrix of substrates combined with the high-dimensional reaction conditions renders exhaustive experimental impractical~\cite{angello2022closed}. 
For decades, chemists have focused on building reliable and convenient computer-aided synthesis planning (CASP) tools to facilitate chemical synthesis~\cite{mikulak2020computational}. Coley et al. built a multiway classification model based on a two-step graph convolutional network (GCN) for the reaction prediction task~\cite{coley2017prediction, coley2019graph}. Nam et al. proposed the first sequence-to-sequence model for forward prediction using the SMILES representations of molecules~\cite{nam2016linking}. Inspired by the attention-based transformer model~\cite{vaswani2017attention}, Schwaller et al. proposed molecular transformers~\cite{schwaller2019molecular, ding2024exploring}, which were applied in forward prediction and reaction condition prediction tasks~\cite{schwaller2019molecular, andronov2023reagent}. Wang et al. reported the application of reinforcement learning models to identify generally applicable conditions~\cite{wang2024identifying}.

% \begin{figure}[h]
% \begin{center}
% %\framebox[4.0in]{$\;$}
% \fbox{\rule[-.5cm]{0cm}{4cm} \rule[-.5cm]{4cm}{0cm}}
% \end{center}
% \caption{Sample figure caption.}
% \end{figure}

Chemical reaction condition prediction tasks aim to identify catalysts, reagents, solvents, or other conditions for a specific reaction. The exploration of a suitable condition is crucial in synthetic chemistry, as it dictates the expected outcomes, including reaction yields and rates~\cite{schnitzer2024machine}. Gao et al. developed a neural network model to predict the chemical context as well as the temperature for any particular organic reaction~\cite{gao2018using}; Maser et al. proposed a machine-learned ranking model to predict the set of conditions used in a reaction as a binary vector~\cite{maser2021multilabel}; Wang et al. proposed Parrot, a powerful and interpretable transformer-based model for the prediction of reaction condition~\cite{wang2023generic}; In the meantime, in order to enhance the representation of reactions, Qian et al.~\cite{qian2023predictive} designed TextReact, which introduced relevant corpus retrieved from literature to enhance the molecular representation of the reaction based on SMILES. Nevertheless, these methods rely on manual feature selection by experts' knowledge and lack a general prediction model with powerful reaction representation.

% LLM for chemistry
Nowadays, the emergence of generative pre-trained transformer-based large language models (LLMs), typified by GPT-4, has triggered keen interest in leveraging such techniques to tackle chemistry challenges~\cite{baum2021artificial, openai2023gpt}. Several works focus on chemical agents for the exploration of chemical conditions. Boiko et al. ~\cite{boiko2023autonomous} proposed a GPT-4 driven scientific agent system to plan and perform complex experiments, which accelerates reaction condition screening and experimental automation in chemistry; Bran et al. developed ChemCrow, which augmented LLMs with chem-expert-designed tools~\cite{m2024augmenting}; 
However, for tasks demanding a precise understanding of molecular SMILES representation, such as reaction prediction, and retrosynthesis, LLMs exhibited a less competitive performance than traditional machine learning baselines~\cite{guo2023can}. Partially, the reason is that, without an in-depth understanding of the SMILES strings and the reaction process that transforms reactants into products, it will be difficult for LLMs to generate accurate responses.
\section{Methods} \label{sec:methods}
% \vspace{-3pt}
% Predefined instruction prompts augmented by corpus and reaction representation are integrated as inputs to LLMs to predict answers.
% A negative log likelihood loss between the generated answer and groundtruth answer is calculated. The modality projection is trained by minimizing this loss.
\vspace{-3pt}
\subsection{Problem Setup}
\vspace{-2pt}
For a task of reaction condition prediction, 
% we define the $X$ as the SMILES input of the chemical reaction $R$, $G$ as the graph representations of reactions, and $T$ as the reaction corpus. The output $Y$ as a list of reaction conditions including the catalyst, solvent, and reagent. Thus, we define the prediction model $\mathcal{F}$, i.e., $Y = \mathcal{F}(X, G, T)$.
we define these symbols for following clarification: (1) reaction SMILES input $R$, reaction SMILES representation encoded by encoder $R_{f}$, and reaction tokens $X$; (2) graph representation of a reaction $G_{f}$ and graph tokens $G$; (3) text corpus of similar reactions $T$, and text tokens $W$; (4) reaction conditions $Y$, which includes the catalyst, solvent, and reagent. We define $\mathcal{F}$ as a condition prediction function, then we obtain:
\begin{equation}
    Y = \mathcal{F}(X, G, T)
\end{equation}

\subsection{Model Structure}
An overview of Chemma-RC is illustrated in Figure~\ref{fig:framework}. Chemma-RC responds to task-specific questions constructed by instruction prompts such as \textit{``which solvents would you recommend?''}, and generate answers as reaction conditions prediction.  Firstly, it learns a unified reaction representation from SMILES, graphs, and corpus. Subsequently, the tunable modality alignment transforms the graph and SMILES embeddings into language tokens compatible with the LLM space. Finally, Chemma-RC generates the SMILES of reagents, solvents, and catalysts as predicted results. 

% Specifically, we freeze the parameters for the reaction and graph modules ((b) and (c) in Figure~\ref{fig:framework}A) and fine-tune the parameters of the last five layers in LLM.

% We develop two distinct types of prediction modules, a classification module and a generation module for Chemma-RC to enhance its compatibility with different chemical reaction conditions. Prediction modules are used to generate probability distributions over potential tokens, and we define two types of loss for this:
To formalize this, let $W=\left\{w_1, w_2, \ldots, w_L\right\}$ be a sequence of word tokens (denoted as \texttt{<Corpus>} in Figure~\ref{fig:prompt}) of length $L$ from the vocabulary $\mathcal{W}$, reaction tokens \texttt{<Reaction>} $X=\left\{x_1, x_2, \ldots, x_k\right\}$ of length $k$, and graph tokens \texttt{<Graph>} $G=\left\{g_1, g_2, \ldots, g_m\right\}$ of length $m$, LLMs parameterized by $\theta$ decompose the joint distribution as $p_{\theta}(W)=\prod_{i=1}^L p_{\theta}\left(w_i \mid W_{<i}\right)$, where $W_{<i}$ represents the tokens preceding the $i$-th position. These learnable reaction and graph tokens, along with text tokens, are then input into the LLM to predict chemical reaction conditions, as shown in the equation~\ref{eq1}:
\begin{equation} \label{eq1}
    \mathcal{L}_{\mathrm{LM}}=\sum_i-\log p_{\theta}\left(w_i \mid W_{<i}, X, G\right)
\end{equation}
 
\subsection{Modality Alignment}

In Figure~\ref{fig:framework}B, we introduce an alignment module designed to facilitate cross-modal representation learning among three distinct modalities. This module leverages latent tokens derived from graph and SMILES embeddings, aligning them with corresponding text-based tokens. To achieve this alignment, we utilize two transformer-based Perceiver modules~\cite{jaegle2021perceiver}, which project the graph and SMILES representations into a shared semantic space compatible with LLMs. Although these Perceiver modules share an identical architecture, they are parameterized independently. The pseudo-code for the modality projection process is detailed in Appendix~\ref{app: alignment}.

To optimize this alignment, we adopt a contrastive learning loss approach. Specifically, an InfoNCE loss, described in equation~\ref{eq:cl}, is designed to minimize the embedding distance between the textual modality and its corresponding graph and SMILES representations for the same reaction ($\mathcal{L}_{\text{text-graph}}$ and $\mathcal{L}_{\text{text-SMILES}}$). Simultaneously, it maximizes the distance between graph and SMILES representations of different reactions ($\mathcal{L}_{\text{graph-SMILES}}$). This dual application of the InfoNCE loss effectively aligns text-graph and text-SMILES pairs from the same reaction while contrasting graph-SMILES pairs from different reactions.

% \vspace{-3pt}
% \begin{align} \label{eq:cl}
% \mathcal{L}_{\text{text-graph}} = 
% - \frac{1}{2} \, \mathbb{E}_{\mathbf{x}_g, \mathbf{x}_t}
% \Bigg[
% & \log \frac{\exp\left(E(\mathbf{x}_g, \mathbf{x}_t)\right)}
% {\exp\left(E(\mathbf{x}_g, \mathbf{x}_g)\right) + \sum_{\mathbf{x}_{t'} \neq \mathbf{x}_t} \exp\left(E(\mathbf{x}_g, \mathbf{x}_{t'})\right)} \nonumber \\
% +\, & \log \frac{\exp\left(E(\mathbf{x}_g, \mathbf{x}_t)\right)}
% {\exp\left(E(\mathbf{x}_g, \mathbf{x}_t)\right) + \sum_{\mathbf{x}_{g'} \neq \mathbf{x}_g} \exp\left(E(\mathbf{x}_{g'}, \mathbf{x}_t)\right)}
% \Bigg]
% \end{align}

% \begin{figure}[htb]
% \centering
% \resizebox{\linewidth}{!}{
% \begin{minipage}{\linewidth}
% \begin{align}
% \scriptsize
% \begin{equation}
% \scalebox{1.}
% {\mathcal{L}_{\text{text-graph}} = 
% & - \frac{1}{2} \, \mathbb{E}_{\mathbf{x}_g, \mathbf{x}_t} \Bigg[ \;
%  \log \frac{
%     \exp\left(E(\mathbf{x}_g, \mathbf{x}_t)\right)}
%     {
%     \exp\left(E(\mathbf{x}_g, \mathbf{x}_g)\right) + 
%     \sum\limits_{\mathbf{x}_{t'} \neq \mathbf{x}_t} \exp\left(E(\mathbf{x}_g, \mathbf{x}_{t'})\right)
%     } \nonumber \\
% +\, &
% \log \frac{
%     \exp\left(E(\mathbf{x}_g, \mathbf{x}_t)\right)}
%     {
%     \exp\left(E(\mathbf{x}_g, \mathbf{x}_t)\right) + 
%     \sum\limits_{\mathbf{x}_{g'} \neq \mathbf{x}_g} \exp\left(E(\mathbf{x}_{g'}, \mathbf{x}_t)\right)
%     }
% \; \Bigg]
% }
% \label{eq:cl}
% % \end{align}
% \end{equation}
% \end{minipage}
% }
% % \caption{Contrastive loss between text and graph representations.}
% \end{figure}

\begin{figure}[ht]
\centering
\scalebox{0.9}{ % 调节缩放比例
\begin{minipage}{\linewidth}
\begin{align}
\mathcal{L}_{\text{text-graph}} = 
& - \frac{1}{2} \, \mathbb{E}_{\mathbf{x}_g, \mathbf{x}_t} \Bigg[ \;
 \log \frac{
    \exp\left(E(\mathbf{x}_g, \mathbf{x}_t)\right)}
    {
    \exp\left(E(\mathbf{x}_g, \mathbf{x}_t)\right) + 
    \sum\limits_{\mathbf{x}_{t'} \neq \mathbf{x}_t} \exp\left(E(\mathbf{x}_g, \mathbf{x}_{t'})\right)
    } \nonumber \\
+\, &
\log \frac{
    \exp\left(E(\mathbf{x}_g, \mathbf{x}_t)\right)}
    {
    \exp\left(E(\mathbf{x}_g, \mathbf{x}_t)\right) + 
    \sum\limits_{\mathbf{x}_{g'} \neq \mathbf{x}_g} \exp\left(E(\mathbf{x}_{g'}, \mathbf{x}_t)\right)
    }
\; \Bigg]
\label{eq:cl}
\end{align}
\end{minipage}
}
\end{figure}

where $\mathbf{x}_{\mathrm{t}}$ and $\mathbf{x}_{\mathrm{g}}$ form the text-graph pair for each reaction, and $\mathbf{x}_{g^{\prime}}$ and $\mathbf{x}_{t^{\prime}}$ are the negative samples randomly sampled from the noise distribution, which we use the empirical data distribution. $E(\cdot)$ is the dot product function on the jointly learned space, that is, $E\left(\mathbf{x}_{\mathrm{g}}, \mathbf{x}_{\mathrm{t}}\right)=\left\langle p_{\mathrm{g}} \circ f_{\mathrm{g}}\left(\mathbf{x}_{\mathrm{g}}\right), p_{\mathrm{t}} \circ f_{\mathrm{t}}\left(\mathbf{x}_{\mathrm{t}}\right)\right\rangle$, where $\circ$ is the function composition.

The final alignment loss for the module can be computed as a weighted sum of all contrastive pairs presented in Equation~\ref{eq:cl_final}, with each term calculated using the InfoNCE loss presented in Equation~\ref{eq:cl}.
\begin{equation} \label{eq:cl_final}
\mathcal{L}_{\text{align}} = 
\frac{1}{3} \left(
\mathcal{L}_{\text{text-graph}} + 
\mathcal{L}_{\text{text-SMILES}} + 
\mathcal{L}_{\text{graph-SMILES}}
\right)
\end{equation}

We propose a two-stage training strategy for Chemma-RC, consisting of supervised fine-tuning (SFT) followed by post fine-tuning, as detailed in Section~\ref{sec: end-to-end training}. Therefore, for the first training stage, the final loss for training Chemma-RC is the integration of next token prediction loss ($\mathcal{L}_{\mathrm{LM}}$) and alignment loss ($ \mathcal{L}_{\mathrm{align}}$), which is illustrated in equation~\ref{eq:all}:

\begin{equation} \label{eq:all}
\mathcal{L}_{\text{final}} = 
\mathcal{L}_{\mathrm{LM}} + 
\mathcal{L}_{\mathrm{align}}
\end{equation}
\subsection{Ranking Enhancement with Feedback (REF)}
Given our focus on the practical applications in synthetic chemistry, we also want to underscore the significance of accurately identifying reaction conditions that lead to high-yield outcomes. Large chemical models such as ChemDFM~\cite{zhao2025developing} commonly use top 50\% accuracy to evaluate ligand prediction performance. This metric is well-suited for high-throughput experimentation (HTE) datasets, where the top-ranked half of ligands generally correspond to satisfactory reaction outcomes. However, when models trained under this paradigm are transferred to literature-based datasets for one-shot condition prediction, they often fail to generate suitable reaction conditions that lead to high yields.

To address this, as depicted in Figure~\ref{fig:framework}C, we design the ranking enhancement module with feedback. This component is the second stage phase of training, as the detailed training process is discussed in Section~\ref{sec: end-to-end training}. It learns the preferences among condition candidates, thereby enabling the model to predict conditions that are optimized for each specific target substrate. Specifically, it provides a prediction score not only for the ground truth preference condition but also for a set of candidates. During post-fine-tuning, we utilize High-Throughput Experimentation (HTE) data for training, which are predefined and sorted according to their yields of reaction outcomes. The objective of this fine-tuning is to learn a ranking function that assigns higher prediction scores to condition candidates of higher performance. We employ a ranking loss that penalizes the model when it fails to rank high-yield candidates above lower-yield ones. The ranking loss is defined as follows:
\vspace{-2pt}
\begin{equation}
    \mathcal{L}_{\text {Ranking}}=\sum_{i=1}^{n-1} \max \left(0, \Delta_i-\operatorname{score}\left(C_i\right)+\operatorname{score}\left(C_{i+1}\right)\right)
\end{equation}

where $\mathcal{L}_{\text {Ranking}}$ is the ranking loss, $n$ is the number of candidates, $\Delta_i$ is the allowed margin between the scores of the $i$-th candidate $C_{i}$ and the $(i+1)$-th candidate $C_{i+1}$, and $\operatorname{score}(C_i)$ is the model’s prediction score for the $i$-th candidate. The loss function encourages the model to learn that the score of the $i$-th candidate should be at least $\Delta_i$ higher than the score of the $(i+1)$-th candidate. If the predictions do not meet this criterion, the loss is non-zero and the model is penalized.

\subsection{End-to-End Model Fine-tuning} \label{sec: end-to-end training}
\textbf{Supervised Fine-Tuning (SFT)}: we employ multimodal SFT to integrate the base LLM with other modules in Chemma-RC. In this process, we freeze the parameters of the reaction and graph encoders depicted in Figure~\ref{fig:framework}A, and focus on fine-tuning the parameters of the LLM, denoted as $\theta$ in Equation~\ref{eq1}, as well as the projection layers for the query tokens $X$ and $G$. The projection layers for the graph and reaction encoders are designed to share the same architecture, although they are parameterized independently. The optimization can be performed end-to-end using Equation~\ref{eq1}.  This fine-tuning stage aligns the LLM with domain-specific models. To preserve the generality of the base LLM, and improve training efficiency, we utilize parameter-efficient LoRA for SFT.

\textbf{Ranking Enhancement with Feedback (REF)}: The REF stage is the second phase of training, following the initial SFT stage. In this phase, we focus exclusively on optimizing the ranking loss, while keeping the parameters fine-tuned during SFT frozen. This approach allows the ranking enhancement module to learn preferences among different reaction condition candidates, improving the model's ability to predict high-yield outcomes for specific target substrates.

\vspace{-3pt}
\subsection{Construction of Multimodal Instruction Datasets}
\begin{figure}[htbp]
  \centering
  \includegraphics[width=1\linewidth]{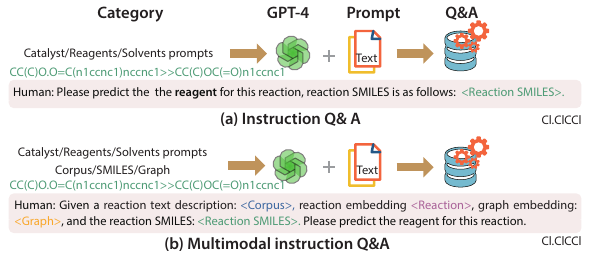}
  \caption{Construction of multimodal instruction datasets. \textbf{(a)} Traditional instruction prompts for supervised fine-tuning; \textbf{(b)} Our proposed text-augmented multimodal instruction Q\&A datasets.}
  \label{fig:prompt}
  \vspace{-10pt}
\end{figure}

\vspace{-2pt}
% Large language model, like ChatGPT, supports prompt engineering, which is an instruction phrase provided to the language model to generate a more reasonable response.
Instruction prompt datasets refer to format structured or unstructured data as natural language instructions so that LLMs can respond properly~\cite{reynolds2021prompt, wang-etal-2023-self-instruct}. 
Here, we introduce a tailored format of instruction-style prompts to facilitate multimodal SFT, shown in Figure~\ref{fig:prompt}. In contrast to standard prompts used in common SFT in Figure~\ref{fig:prompt}(a), we incorporate additional textual indicators and modality-specific tokens (Figure~\ref{fig:prompt}(b)) to represent multimodal chemical data.
%%%%%%%%%%%%%%%%%%%%%%%%%%%%%%%%%%%%%%%%%%%%%%%%%%%%%%%%%
%%%%%%%%%%%%%%%%%%%%%%%%%%%%%%%%%%%%%%%%%%%%%%%%%%%%%%%%%

To be specific, given a reaction, we retrieve a relevant corpus—a paragraph containing contextual information that closely resembles the reaction—and populate the \texttt{<Corpus>} placeholder with this data. Next, the reaction is converted into its corresponding SMILES format, and inserted into the specific token placeholder \texttt{<Reaction SMILES>}. Finally, we introduce two additional token placeholders, \texttt{<Reaction>} and \texttt{<Graph>}, for reaction SMILES and graph representations, respectively. To enhance the diversity of the dataset, we further leverage GPT-4 to generate a wide range of question prompts. These prompts are based on predefined templates, examples of which are provided in Table~\ref{tab:question prompt}.

\section{Experiments and Results} \label{sec:experiment}

\subsection{Data}
We evaluate on two benchmark datasets, USPTO-Condition and USPTO\_500MT\_Condition, respectively. The details of data description are presented in Appendix~\ref{app: data description} and Table~\ref{tab:data format}. The visualization of data distribution is depicted in Figure~\ref{fig:dataset}. As shown in Table~\ref{tab:data format}, the reaction conditions in the USPTO-Condition dataset are divided into \textbf{five distinct categories}, such as \textit{solvent 1}, \textit{reagent 1}, etc, and presented in a fixed order, which results in a more structured prediction task. In contrast, all reaction conditions in USPTO\_500MT\_Condition dataset are \textbf{a single dot-concatenated string}, posing a greater challenge due to the need to generate an unstructured sequence with correct formatting and semantics.

\subsection{Experiment Setup}
In our work, the reaction encoder is implemented based on Wang et al.~\cite{wang2023generic}. A pre-trained graph model proposed by~\cite{schlichtkrull2018modeling} encodes the molecules in the reaction. We utilize LLaMA-2~\cite{llama-2} as a text decoder. Each reaction has a corresponding similar corpus, a paragraph describing a chemical reaction with an average length of 190 tokens.
% For the reaction condition prediction task, Chemma-RC generates the reaction conditions in a specific order: catalyst, solvent 1, solvent 2, reagent 1, and reagent 2, which follows the classification policy proposed by Gao et al.~\cite{gao2018using}.
During the training process, we freeze the weight parameters of GCN and the reaction encoder. The modality alignment and part layers of LLaMA-2 are trainable. We utilize parameter-efficient LoRA for SFT, and the trainable parameters constitute approximately \textbf{0.3 billion} out of the total \textbf{7 billion} parameters. The multimodal SFT process is conducted with a batch size of 16 for fewer than 6 epochs over 48 hours, utilizing a GPU configuration of 2 $\times$ 48 GB NVIDIA A6000 GPUs. Inference is performed on a single 80 GB NVIDIA A800 GPU. Detailed training configuration is shown in Appendix~\ref{app: training settings}. 

% Given a SMILES of reaction, pair-wised Q\&A instruction prompts, and its similar corpus, reaction representation encoded from GCN and transformer-based model are then fed into the modality projection to produce additional tokens. The additional tokens and text-augmented tokens are aggregated as the input of the LLaMA-2 to generate a final answer.

\subsection{Performance Comparison}
We conduct a systematic evaluation to demonstrate Chemma-RC's superior performance for reaction condition prediction. Compared baseline methods include rxnfp LSTM~\cite{gao2018using}, Reaction GCNN~\cite{maser2021multilabel}, TextReact~\cite{qian2023predictive}, and Reagent Transformer~\cite{andronov2023reagent}. The detailed introduction of these methods is presented in Appendix~\ref{app: model performance}.

%%%%%%%%%%%%%%%%%%%%%%%%%%%%%%%%%%%%%%%%%%%%%%%%%%%%%%%%

\begin{table*}[htbp]
    
    \centering
    \caption{Accuracy results for reaction condition prediction on USPTO-Condition dataset. The best performance is in \textbf{bold}.}
    % \vspace{-5pt}
    \resizebox{1\linewidth}{!}{
    \begin{tabular}{l c ccc c ccc c ccc c ccc c ccc}
    \toprule
    \multicolumn{1}{c}{\multirow{3}{*}{\textbf{Model}}} & & \multicolumn{19}{c}{\textbf{Top-$k$ Accuracy (\%)}} \\
    \cmidrule{3-21}
         & & \multicolumn{3}{c}{\textbf{Catalyst}} & & \multicolumn{3}{c}{\textbf{Solvent 1}} & & \multicolumn{3}{c}{\textbf{Solvent 2}} & & \multicolumn{3}{c}{\textbf{Reagent 1}} & & \multicolumn{3}{c}{\textbf{Reagent 2}} \\
         \cmidrule{3-5} \cmidrule{7-9} \cmidrule{11-13} \cmidrule{15-17} \cmidrule{19-21}
         & & 1 & 3 & 5 && 1 & 3 & 5 && 1 & 3& 5 && 1 & 3& 5&& 1 & 3 & 5\\
         \midrule
         % rxnfp LSTM & & 91.6 & 94.1 & 95.2 &&48.3 & 64.4 & 70.2 && 81.4 & 83.4 & 84.6 && 48.2 & 64.4 & 70.8 && 76.5 & 84.1& 86.4\\
         rxnfp LSTM & & 92.2 & 92.2 & 92.2 && 50.2 & 66.4 & 70.6 && 81.3 & 83.7 & 84.6 && 49.7 & 66.0 & 74.0 && 76.2 & 84.1 & 86.6\\
         % Parrot && 89.9 & 96.4 & 97.7 & & 35.2 & 60.9 & 72.2 && 81.2 & 93.7 & 96.7 && 40.4 & 62.3 & 71.7 && \textbf{80.6} & 90.6 & 93.6 \\
         Parrot && 92.4 & 92.4 & 92.4 & & 49.3 & 67.7 & 72.3 && 80.7 & 84.2 & 85.1 && 49.6 & 67.3 & 75.7 && 76.5 & 84.1 & 87.2 \\
         % TextReact$_s$ && 90.3 & 93.4 & 94.8 && 44.6 & 62.9 & 70.8 && 77.5 & 82.9 & 85.2 && 43.1 & 60.3 & 68.1 && 75.1 & 82.0 & 85.3 \\
         % TextReact$_s$ && 92.1 & 98.0 & 99.1 && 51.4 & 68.5 & 79.3 && 81.6 & 93.4 & 96.9 && 51.1 & 69.6 & 79.1 && 77.9 & 91.1 & 94.9 \\
         TextReact$_s$ && 92.4 & 95.3 & 96.3 && 51.7 & 65.5 & 71.7 && 79.8 & 87.7 & 89.8 && 51.9 & 68.7 & 75.1 && 75.8 & 86.7 & 89.7 \\
         % Reacon && 92.4 & - & - && 50.4 & - & - && 81.6 & - & - && 50.0 & - & - && 78.0 & - & -\\
         % Reaction Graph && \textbf{93.2} & 92.3 & 93.2 && 54.3 & 69.3 & 72.7 && 80.8 & 85.6 & 86.5 && \textbf{53.4} & 69.8 & 77.1 && 76.3 & 86.6 & 89.3 \\
         \rowcolor{mygray}
         Chemma-RC &&\textbf{92.7} & \textbf{98.6} & \textbf{99.2} && \textbf{54.6} & \textbf{76.4} & \textbf{84.8} && \textbf{81.8} & \textbf{94.8} & \textbf{97.6} && \textbf{53.4} & \textbf{75.8} & \textbf{83.9} && \textbf{78.7} & \textbf{93.2} & \textbf{96.2}\\
         % Chemma-RC$_g$ &&98.7 &99.9 &99.9 && 92.9 & 98.4 & 99.0 && 92.0 & 99.2 & 99.7 && 87.4 & 97.0 & 98.3 && 92.4& 99.3& 99.6\\
         \bottomrule
    \end{tabular}
    }
    \label{tab:performance-condition}
    % \vspace{-5pt}
\end{table*}

For the USPTO-Condition dataset, we calculate top-$k$ accuracy with a strict matching policy. All SMILES from the prediction results are canonicalized to ensure consistent comparison. As depicted in Table~\ref{tab:performance-condition}, TextReact$_s$ refers that we utilize \textit{similar text}~\cite{qian2023predictive} paired with the corresponding reaction for training. To avoid label leak issues, we do not use \textit{gold text} mentioned in his work for training or testing. Thanks to the work of Qian et al., we retrieve the most semantically relevant corpus entries from literature or patents for each reaction. These retrieved corpus are integrated with reactions to construct Q\&A instruction datasets for multimodal SFT.
% \begin{wrapfigure}{l}{0.48\textwidth}
% \begin{table}[htbp]
% % {l}{0.48\textwidth}
%     \centering
%     \scriptsize
%     % \vspace{-10pt}
%     \captionof{table}{Results of reaction condition prediction on USPTO\_500MT\_Condition dataset. The best performance is in \textbf{bold}.\label{tab:performance-500mt}}
%     \resizebox{0.9\linewidth}{!}{\begin{tabular}{l c cccc}
%     \toprule
%        \multicolumn{1}{c}{\multirow{2}{*}{\textbf{Model}}}  &&  \multicolumn{4}{c}{\textbf{Top-$k$ Accuracy (\%)}} \\
%        \cline{3-6}
%          & & 1 & 3 & 5 & 10 \\
%          \midrule
%          Reagent Transformer && 17.5 & 27.5 & 31.6 & 35.6 \\
%          % \toprule
%          Reaction GCNN && 16.1 & 27.5 & 33.0 & 40.2 \\
%          % \midrule
%          Parrot && 13.8 & 25.3 & 31.4 & 37.9 \\
%          % \midrule
%          nach0 && 13.1 & - & - & -  \\
         
%          \rowcolor{mygray}
%          \textbf{Chemma-RC} && \textbf{25.9} & \textbf{47.2} & \textbf{67.8} & \textbf{79.2} \\
%          \bottomrule
%     % \vspace{-40pt}
%     \end{tabular}}
% \end{table}
% \end{wrapfigure}

%%%%%%%%%%%%%%%%%%%%%%%%%%%%%%%%%%%%%%%%%%%%%%%%%%%%%%%%%

% Figure~\ref{fig:pf-radar}A illustrates the relative performance enhancement over the rxnfp LSTM method on the USPTO-Condition dataset. The radar plots show top-1 (left), top-3 (middle), and top-5 (right) accuracy improvements for three methods: Parrot (red), TextReact (green), and Chemma-RC (blue). 

The overall performance is summarized in the Table~\ref{tab:performance-condition}. From the results, we can see that Chemma-RC consistently outperforms the baselines across all categories and accuracy levels, yielding a significant improvement of 7\% over TextReact. Specifically, Chemma-RC achieves superior performance compared to other methods, attaining top-1 accuracy of 54.6\% for solvent 1 and 81.8\% for solvent 2 prediction, respectively. We also observe that the observed performance disparity across condition types—such as the significantly higher accuracy for catalyst prediction (92.7\%) compared to solvent 1 prediction (54.6\%). It can be attributed to the inherent distributional differences within the dataset. Specifically, the statistical results in Table~\ref{tab:sparsity-USPTO-Condition} illustrate that the number of distinct catalyst types present in the training data is relatively limited, which leads to the highly consistent usage across reactions. In contrast, the solvent category, particularly solvent 1, exhibits much greater chemical diversity, with a larger number of unique solvents and more varied usage contexts. This increased diversity results in a more challenging prediction task, leading to lower model performance in this category.

% The overall top-1 accuracy of Chemma-RC is 34.1\% higher than that of the Parrot model. 

For the USPTO\_500MT\_Condition dataset, all reaction conditions are a single dot-concatenated string, annotated as \textit{reagents}. All reaction conditions must be generated in a single inference pass and then canonicalized to ensure consistency for evaluation and comparison. In Table~\ref{tab:performance-500mt}, we report two metrics, including top-1 accuracy and partial accuracy. Different from the complete match accuracy that requires an exact match between the predicted and ground-truth conditions, the partial match accuracy focuses more on evaluating whether individual components (e.g., solvent, reagent, or catalyst) are correctly predicted, even if the full sequence is not perfectly matched.
Relative enhanced performance is visualized in Appendix Figure~\ref{fig:pf-radar}. Notably, Chemma-RC significantly outperforms other LLMs and domain-specific chemical models~\cite{andronov2023reagent, livne2024nach0} in Table~\ref{tab:performance-500mt}. In the zero-shot setting, Chemma-RC achieves a top-1 accuracy of 25.9\%, which is significantly higher than the best-performing general-purpose model, ChemDFM~\cite{zhao2024chemdfm}, at 2.0\%. Further, we investigate the distribution of condition numbers combinations in test set, and report both top-1 exact match accuracy and partial accuracy in Table~\ref{tab:reagent-count-performance}. We find that exact match accuracy, as well as precision and recall, increases with the frequency of specific condition number combinations in the dataset, irrespective of the type or quantity of reagents involved. Specifically, for three-condition combinations, which occur 3,258 times in the test set, Chemma-RC achieves a higher partial accuracy of 85.6\%, compared to 56.91\% in the one-condition scenario with 1,622 occurrences. Furthermore,  we select several challenging reactions for detailed discussion, with results presented in Table~\ref{tab:reaction-type-performance}. Our model, Chemma-RC, demonstrates robust performance in predicting complex chemical reactions, such as ring cleavage, achieving an exact match accuracy of 66.13\% across six different condition combinations. In summary, despite single-condition samples being simpler in structure, the limited occurrence provides fewer learning signals, which can negatively impact generalization. Conversely, frequent multi-condition examples offer richer and more consistent patterns, leading to better model performance, especially in partial accuracy. 

%%%%%%%%%%%%%%%%%%%%%%%%%%%%%%%%%%%%%%%%%%%%%%%%%%%%%%%%%
\begin{table*}[htbp]
\centering
\caption{Comparison of model performance on USPTO\_500MT\_Condition dataset between general-purpose LLMs and domain-specific chemical models, respectively. General-purpose LLMs are tested under three settings: zero-shot, one-shot, and five-shot. Error analysis is reported, and the best results are in bold font.}
\scriptsize
\resizebox{0.7\linewidth}{!}{
{\begin{tabular}{l|cccccc}
\toprule
\textbf{Method} & \textbf{Top-1 exact acc. (\%)} & \textbf{Partial acc. (\%)} & \textbf{Recall (\%)} & \textbf{Precision (\%)} \\
\midrule

\multicolumn{5}{l}{\textit{General-purpose LLMs, 100 examples are randomly sampled for evaluation}} \\
\midrule
\rowcolor{cyan!10}
\multicolumn{5}{l}{\textbf{\textit{Zero-shot performance}}} \\
DeepSeek-V2   & $0.0 \pm 0.019$ & $15.1 \pm 0.018$ & $17.4 \pm 0.018$ & $17.9 \pm 0.034$ \\
GPT-4o        & $1.0 \pm 0.012$ & $13.0 \pm 0.010$ & $7.6 \pm 0.034$ & $12.1 \pm 0.038$ \\
LLaMA3-70B    & $0.0 \pm 0.000$ & $7.0 \pm 0.010$  & $11.4 \pm 0.024$ & $9.1 \pm 0.024$ \\
ChemDFM       & $3.0 \pm 0.029$ & $38.0 \pm 0.024$ & $19.6 \pm 0.012$ & $26.5 \pm 0.028$ \\

\midrule
\rowcolor{cyan!10}
\multicolumn{5}{l}{\textbf{\textit{One-shot performance}}} \\

DeepSeek-V2   & $0.0 \pm 0.019$ & $15.1 \pm 0.018$ & $17.4 \pm 0.018$ & $17.9 \pm 0.034$ \\
GPT-4o        & $1.0 \pm 0.012$ & $13.0 \pm 0.010$ & $7.6 \pm 0.034$  & $12.1 \pm 0.038$ \\
LLaMA3-70B    & $0.0 \pm 0.000$ & $7.0 \pm 0.010$  & $11.4 \pm 0.024$ & $9.1 \pm 0.024$  \\
ChemDFM       & $3.0 \pm 0.029$ & $38.0 \pm 0.024$ & $19.6 \pm 0.012$ & $26.5 \pm 0.028$ \\

\midrule
\rowcolor{cyan!10}
\multicolumn{5}{l}{\textbf{\textit{Five-shot performance}}} \\
DeepSeek-V2   & $1.3 \pm 0.014$ & $15.8 \pm 0.059$ & $17.2 \pm 0.018$ & $25.3 \pm 0.050$ \\
GPT-4o        & $0.0 \pm 0.039$ & $15.3 \pm 0.079$ & $6.2 \pm 0.036$  & $9.8 \pm 0.020$  \\
LLaMA3-70B    & $1.0 \pm 0.010$ & $28.0 \pm 0.013$ & $14.1 \pm 0.024$ & $11.9 \pm 0.024$ \\
ChemDFM       & $1.0 \pm 0.017$ & $21.0 \pm 0.070$ & $11.1 \pm 0.038$ & $14.4 \pm 0.073$ \\

\midrule
\multicolumn{5}{l}{\textit{Domain-specific chemical models, all samples in test sets are selected for evaluation}} \\
\midrule
Reagent Transformer & 17.5 & 27.5 & 31.6 & 35.6 \\
Reaction GCNN & 16.1 & 27.5 & 33.0 & 40.2 \\
Parrot & 13.8 & 25.3 & 31.4 & 37.9 \\
nach0 & 13.1 & - & - & -  \\

\midrule
\textbf{Chemma-RC (zero-shot)} & \textbf{25.9} & \textbf{69.7} & \textbf{67.9} & \textbf{79.3}  \\

\bottomrule
\end{tabular}}
}
\label{tab:performance-500mt}
\end{table*}

\vspace{-3pt}
\subsection{Ablation Study} \label{sec:ablation}
\vspace{-2pt}
\subsubsection{Model structure}
\begin{figure*}[htbp]
  \centering
  \includegraphics[width=0.88\linewidth]{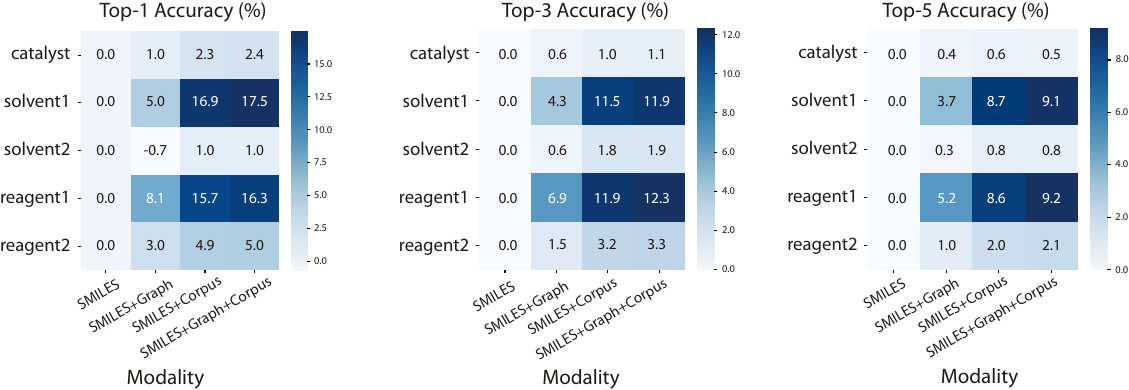}
  \caption{Heatmap visualization of performance enhancements of Chemma-RC on the USPTO-Condition dataset.}
  \label{fig:ablation-incremental}
  % \vspace{-10pt}
\end{figure*}

% graph 效果不明显的原因是 他本身和smiles表征接近。但相对于对于文本的表征而言，提升不会很大。
% experiment description
Here, we conduct an ablation study to examine the inherent effect of different modalities on the performance of Chemma-RC. Specifically, we evaluate the performance under the different combinations of mono-domain, including SMILES, graph, and corpus, on the USPTO-Condition dataset. The results are reported in Table~\ref{tab:mono-data}, and the heatmap visualization illustrating the relative performance improvements contributed by different modalities is presented in Figure~\ref{fig:ablation-incremental}. The results clearly indicate that different mono-domain inputs contribute unevenly to overall performance. For the prediction of solvent 1, which emerges as the most challenging condition type (in Table~\ref{tab:performance-condition}), the model enhanced with SMILES modality (first row) outperforms the models trained solely on graph-based modality (second row) and corpus data (third row), achieving 21.8\% and 23.0\% higher top-1 accuracy, respectively. 

\begin{table*}[htbp] 
 \centering
 
  \renewcommand{\arraystretch}{1.1}
  \caption{Performance evaluation of Chemma-RC under different combinations of mono-domain data on the USPTO-Condition Dataset.}
  % \vspace{-8pt}
  % \vspace{-3pt}
  \resizebox{1\linewidth}{!}{\begin{tabular}{ccccccccccccccccccccccc}
  % {y{50}x{20}x{20}x{5}x{20}x{20}x{20}x{5}x{20}x{20}x{20}x{5}x{20}x{20}x{20}x{5}x{20}x{20}x{20}}
  \toprule
  \multirow{3}{*}{\textbf{SMILES}} & \multirow{3}{*}{\textbf{Graph}}  & \multirow{3}{*}{\textbf{Corpus}} & \multicolumn{19}{c}{\textbf{Top-$k$ Accuracy (\%)}}\\
  \cmidrule{5-23}
  && & & \multicolumn{3}{c}{\textbf{Catalyst}} & & \multicolumn{3}{c}{\textbf{Solvent 1}} & & \multicolumn{3}{c}{\textbf{Solvent 2}} & &\multicolumn{3}{c}{\textbf{Reagent 1}}& &\multicolumn{3}{c}{\textbf{Reagent 2}}\\
  \cmidrule{5-7}
  \cmidrule{9-11}
  \cmidrule{13-15}
  \cmidrule{17-19}
    \cmidrule{21-23}
  & & & & \textbf{1} & \textbf{3}   & \textbf{5}  &    & \textbf{1}   & \textbf{3}  & \textbf{5}    &  & \textbf{1}   & \textbf{3}  & \textbf{5}    &  & \textbf{1}   & \textbf{3} & \textbf{5} &  & \textbf{1}   & \textbf{3} & \textbf{5}\\
  \midrule
    % \multirow{2}{*}{SSD-based}
  \checkmark & \ding{55}   & \ding{55}   &   & 90.3 & 97.5  & 98.7    &  & 37.1  & 64.5 & 75.7    &  & 80.8 & 92.9  & 96.8    &  & 37.1 & 63.5 & 74.7    &  & 73.7  & 89.9 & 94.1 \\
  \ding{55} & \checkmark & \ding{55}  &  & 87.1 & 93.3 & 95.5   &  & 15.3 & 40.5 & 58.2  & &  80.7 & 91.9 & 95.5  & & 34.6 & 56.8 & 67.5 & & 75.4 & 86.6 & 90.6\\

  \ding{55} & \ding{55}  & \checkmark   &   & 87.1 & 87.4  & 87.8   &   & 14.1 &26.1  & 44.9  &   & 80.7 & 88.1  & 92   &   & 26.0 &32.1  & 37.3 & & 75.1 & 76.6 & 77.9 \\

  \checkmark & \ding{55}   & \checkmark   &   & 92.6 & 98.5  & \textbf{99.3}    &  & 54.0  & 76.0 & 84.4    &  & \textbf{81.8} & 94.7  & \textbf{97.6}    &  & 52.8 & 75.4  & 83.3    &  & 78.6  & 93.1 & 96.1 \\

  \checkmark & \checkmark & \ding{55} &  & 91.3 & 98.1 & 99.1 & & 42.1 & 68.8 & 79.4 & & 80.1 & 93.5 & 97.1 & & 45.2 & 70.4 & 79.9 && 76.7 & 91.4 & 95.1  \\

\rowcolor{mygray}
  \cellcolor{mygreen}
  {\checkmark} & \cellcolor{mygreen}{\checkmark}  & \cellcolor{mygreen}{\checkmark}   &  & \textbf{92.7}  & \textbf{98.6}  & 99.2     &  &\textbf{54.6}  & \textbf{76.4}  & \textbf{84.8}    &  & \textbf{81.8} & \textbf{94.8}  & \textbf{97.6}    &  & \textbf{53.4} & \textbf{75.8}  & \textbf{83.9}    &  & \textbf{78.7}  & \textbf{93.2} & \textbf{96.2} \\
 \bottomrule
 \end{tabular}}
 \label{tab:mono-data}
 % \vspace{-8pt}
\end{table*}
%%%%%%%%%%%%%%%%%%%%%%%%%%%%%%%%%%%%%%%%%%%%%%%%%%%%%%%%%
%%%%%%%%%%%%%%%%%%%%%%%%%%%%%%%%%%%%%%%%%%%%%%%%%%%%%%%%%

Subsequently, we investigate how chemical mono-domain data combination affects model performance compared to individual types of data (fourth row to sixth row). \textbf{By incorporating a corpus into the model already trained with SMILES representations, we achieve a 16.9\% improvement in solvent 1 top-1 prediction accuracy.} 
However, integrating graph representations into the SMILES-based model results in a 5.0\% improvement in solvent 1 top-1 accuracy. The limited performance gain observed in this setting can be attributed to the inherent similarity between the graph modality and the SMILES representation. Since both modalities encode comparable structural information about the molecule, the removal of one does not significantly impact the model's overall performance. In contrast, the incorporation of textual modality, which introduces complementary semantic and contextual information, has a more substantial effect on representation quality and prediction accuracy.
Although the advantages of incorporating graph modality may not be immediately apparent from aggregated performance metrics, we assert its critical importance for this task. \textbf{In the context of challenging reactions with substrates comprising over 100 atoms, the integration of graph modality has been observed to significantly enhance condition prediction performance, which can be seen in Figure~\ref{fig:case-study-baseline}.} We hypothesize that this improvement arises because the graph modality enables the model to discern subtle differences between complex substrates—nuances that are not adequately captured by SMILES representations alone—thereby facilitating a more accurate prediction of reaction conditions. Therefore, the integration of graph modalities into predictive models will become essential for the actual applications of organic chemistry.

 % 总结

\subsubsection{Out-of-distribution evaluation}\label{sec:data split}
Here, we evaluate the generalization performance of Chemma-RC. We conduct two out-of-distribution (OOD) experiments for evaluation. Firstly, inspired by the work proposed by Qian et al.,~\cite{qian2023predictive}, we also evaluate the out-of-distribution (OOD) performance of Chemma-RC across different dataset splitting strategies on the USPTO-Condition dataset. Secondly, we employ Chemma-RC trained on the USPTO\_500MT\\ \_Condition to test on the USPTO-Condition. Results are presented in Table~\ref{tab:cross-data} and Table~\ref{tab: data split}. We consider both random split (RS) and time-based split (TS) to further assess the model's robustness and generalization ability. Random split (RS) setup follows the original data split of the USPTO-Condition dataset. The second setup--time split is more challenging~\cite{coley2017prediction, gao2018using}, where the dataset is partitioned based on the publication year of patents. We train the model with historical data from older patents and test its performance on the data from newer patents. This temporal division introduces a substantial domain shift, as the difference between reactions in new patents and previous ones. 

\begin{table*}[h]
   \centering
   % \vspace{-8pt}
   % \scriptsize
   \caption{Evaluation performance under different data split strategies for reaction condition prediction. RS: random split; TS: time split.}  
   \resizebox{0.7\linewidth}{!}{
    \begin{tabular}{lccccccccc}
        \toprule
            & \multicolumn{4}{c}{\textbf{Random split}} && \multicolumn{4}{c}{\textbf{Time split}} \\\cmidrule{2-5} \cmidrule{7-10}
            & \textbf{Top-1} & \textbf{Top-3} & \textbf{Top-10} & \textbf{Top-15} && \textbf{Top-1} & \textbf{Top-3} & \textbf{Top-10} & \textbf{Top-15} \\\midrule
        rxnfp LSTM
            & 20.5 & 30.7 & 41.7 & 45.3 && 15.2 & 26.2 & 40.7 & 45.4 \\
        rxnfp retrieval 
            & 27.2 & 37.5 & 47.9 & 51.1 && 7.8  & 15.2 & 27.3 & 31.5 \\
        Transformer 
            & 30.0 & 43.8 & 56.7 & 60.5 && 18.7 & 31.8 & 47.6 & 52.7 \\
        ChemBERTa
            & 30.3 & 44.7 & 58.0 & 62.0 && 18.7 & 31.9 & 47.6 & 52.8 \\ 
            %\midrule
        TextReact(gr)
            % & \\
        % $\ \cdot$ full corpus 
            & 47.2 & 59.9 & 65.0 & 71.4 && 
            36.3 & 50.4 & 56.2 & 63.8\\
        \rowcolor{mygray}
        Chemma-RC
            &  \textbf{72.3} & \textbf{87.8} & \textbf{92.4} & \textbf{96.5} && \textbf{69.6} & \textbf{86.7} & \textbf{91.7} & \textbf{96.2} \\
            
        % $\ \cdot$ gold-removed
        %     & 47.2 & 59.9 & 69.0 & 72.5 && 36.3 & 50.4 & 63.8 & 67.9 \\
        % $\ \cdot$ TS corpus
        %     & ---  & ---  & ---  & ---  && 21.1 & 35.2 & 51.0 & 56.1 \\
        \bottomrule
    \end{tabular}
    % }
   } \label{tab: data split}
   \vspace{-8pt}
\end{table*}

In Table~\ref{tab: data split}, we calculate the average accuracy of all different types of conditions, and report average top-1, top-3, top-10, and top-15 accuracy metrics. TextReact (gr) refers to the TextReact model without retrieving gold texts for testing. The results demonstrate that while baseline method such as ChemBERTa~\cite{chithrananda2020chemberta} achieves moderate success, they fall short in capturing the full complexity of condition prediction. In contrast, Chemma-RC significantly outperforms all baseline methods across both RS and TS settings. Notably, despite being trained only on historical data, Chemma-RC achieves a top-1 (TS) accuracy of 69.6\%. Further, the results of cross-dataset evaluation are presented in Appendix~\ref{app: general} Table~\ref{tab:cross-data}. This substantial improvement highlights Chemma-RC’s superior ability to leverage multi-modal information and its robustness across different data distributions.

\subsubsection{Ranking enhancement with feedback}

%%%%%%%%%%%%%%%%%%%%%%%%%%%%%%%%%%%%%%%%%%%%%%%%%%%%%%%%%
\begin{figure}[h]
 % \vspace{-10pt}
  \centering
  \includegraphics[width=\linewidth]{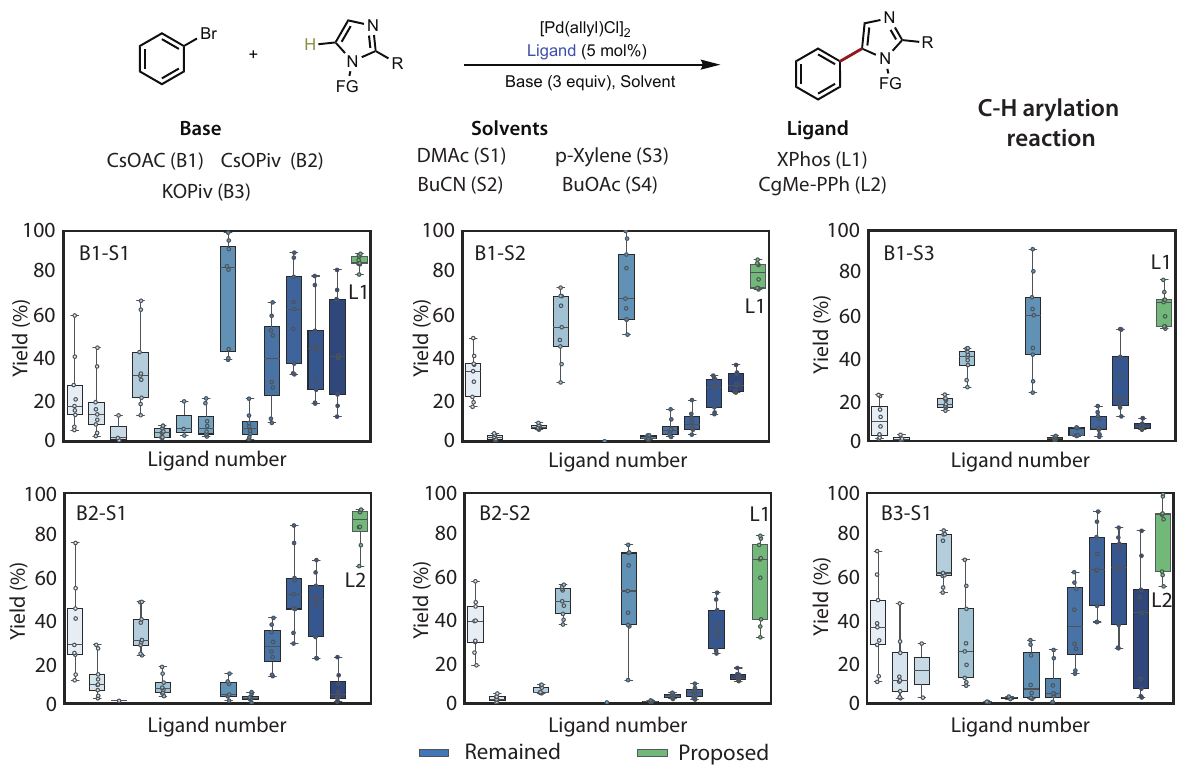}
  \caption{Performance evaluation for identifying optimal ligand on C-H arylation reaction.}
  \label{fig:ligand-recommendation}
   % \vspace{-10pt}
\end{figure}
%%%%%%%%%%%%%%%%%%%%%%%%%%%%%%%%%%%%%%%%%%%%%%%%%%%%%%%%%

%%%%%%%%%%%%%%%%%%%%%%%%%%%%%%%%%%%%%%%%%%%%%%%%%%%%%%%%%

\begin{table*}[t]
  
    \centering
    \caption{Performance evaluation of Chemma-RC under different Modality alignments, the best performance are in bold.}
    % \vspace{-5pt}
    \resizebox{0.9\linewidth}{!}{
    \begin{tabular}{l c ccc c ccc c ccc c ccc c ccc}
    \toprule
    \multicolumn{1}{c}{\multirow{3}{*}{\makecell{\textbf{Projection}\\ \textbf{Layer}}}} & & \multicolumn{19}{c}{\textbf{Top-$k$ Accuracy (\%)}} \\
    \cmidrule{3-21}
         & & \multicolumn{3}{c}{\textbf{Catalyst}} & & \multicolumn{3}{c}{\textbf{Solvent 1}} & & \multicolumn{3}{c}{\textbf{Solvent 2}} & & \multicolumn{3}{c}{\textbf{Reagent 1}} & & \multicolumn{3}{c}{\textbf{Reagent 2}} \\
         \cmidrule{3-5} \cmidrule{7-9} \cmidrule{11-13} \cmidrule{15-17} \cmidrule{19-21}
         & & 1 & 3 & 5 && 1 & 3 & 5 && 1 & 3& 5 && 1 & 3& 5&& 1 & 3 & 5\\
         \midrule
         MLP & & 90.9 &97.8 &98.9 && 51.1 & 73.3 & 82.2 && 81.1 & 93.9 & 97.1 && 47.4 & 71.0 & 79.9 && 77.0& 91.7& 95.2\\
         Reprogramming && 92.1 &98.3 &99.1 && 52.8 & 75.1 & 83.7 && 81.3 & 94.3 & 97.4 && 50.2 & 73.5 & 81.9 && 77.7& 92.5& 95.7 \\
         \rowcolor{mygray}
         Perceiver && \textbf{92.7} &\textbf{98.6} &\textbf{{99.2}} && \textbf{54.6} & \textbf{76.4} & \textbf{84.8} && \textbf{81.8} & \textbf{94.8} & \textbf{97.6} && \textbf{53.4} & \textbf{75.8} & \textbf{83.9} && \textbf{78.7} & \textbf{93.2} & \textbf{96.2}\\
         \bottomrule
    \end{tabular}
    }
    \label{tab:projection ablation}
    % \vspace{-10pt}
\end{table*}

\vspace{-4pt}
% background 当把high-quality candidates rank靠前的时候就奖励，
As we discussed before, we designed a ranking function that assigns higher scores to condition candidates associated with better experimental performance. Here, we curate a Pd-catalysed imidazole C--H arylation~\cite{shields2021bayesian} HTE reaction data from ORD~\cite{coley2017prediction} to evaluate the effectiveness of this module.
% We evaluate the performance on the C--H arylation reaction, which consists of boronic acid derivatives, aryl halide substrates.
The objective is to identify ligands that maximize reaction yield, under the constraint that all other conditions, including bases and solvents, remain fixed within a predefined reaction space. Compared with the top-50\% metric proposed in ~\cite{guo2023can}, we aim to predict the ligands with the highest yields. In Figure~\ref{fig:ligand-recommendation}, the box plot illustrates the yield distribution under different base-solvent-ligand combination of conditions; the box marked in green is the ligand generated by our proposed Chemma-RC. For example, in the left top panel, under the combination of CsOAc and DMAc, Chemma-RC identifies the XPhos ligand. We further evaluate the model proposed by Zhao et al.'s work~\cite{zhao2024chemdfm}, which yields a top-1 accuracy of 38.1\% for ligand selection. In comparison, Chemma-RC achieves a significantly higher accuracy of 93.7\%, representing an improvement of 58.7\%. All results are presented in Appendix Figure~\ref{fig:ligand recommendation1}. In summary, Chemma-RC is capable of generating optimal conditions due to post-fine-tuning by the ranking enhancement module. We hope this technology has the potential to accelerate high-throughput reaction condition screening in the future.

%%%%%%%%%%%%%%%%%%%%%%%%%%%%%%%%%%%%%%%%%%%%%%%%%%%%%%%%%

\subsubsection{Modality alignment}
% background
% By leveraging the strengths of multiple modalities, multimodal LLMs can achieve higher accuracy in a wide range of applications. However, aligning representations among different modalities remains a challenging task. 
In Chemma-RC, the modality alignment module utilizes the Perceiver projection module~\cite{jaegle2021perceiver} to extract latent tokens from both graph and SMILES representations and subsequently aligns these tokens into a text-related language space, as illustrated in Figure~\ref{fig:framework}. Here, we investigate the impact of different projection modules on modality alignment, a component that plays a crucial role in the performance of LMMs. Specifically, we introduce three projection methods for modality alignment, including Perceiver~\cite{jaegle2021perceiver}, Reprogramming~\cite{jin2023time}, and MLP for comparison.

% method
% \begin{enumerate}
%     \item Perceiver~\cite{jaegle2021perceiver}. It leverages an asymmetric attention mechanism to iteratively distill input into a tight latent bottleneck. We apply this module to distill the high-dimensional representation of SMILES and graphs.
%     \item Reprogramming~\cite{jin2023time}. It proposes reprogramming time series with the source data modality along with prompting to unleash the full potential of LLMs as versatile forecasters in standard, few-shot, and zero-shot scenarios. We reproduce this module to align reaction representation with instruction prompts.
%     \item MLP. Representation of each modality is subsequently projected to the space of LLMs with simple linear layers for final prediction.
% \end{enumerate}

% conclusion

% experiment
As depicted in Table~\ref{tab:projection ablation}, the Perceiver module achieves significant gains in the prediction of all condition categories. Compared to reprogramming, it achieves the highest accuracy in all predicted condition categories with an average performance gain of 7.1\%. Specifically, for the solvent 1 prediction, a challenging task, the Perceiver module achieves a top-1 accuracy of 54.6\%, clearly outperforming both MLP (51.1\%) and Reprogramming (52.8\%). This performance demonstrates the Perceiver’s robustness and effectiveness in capturing complex cross-modal relationships, making it a strong candidate for accurate and reliable reaction condition prediction.
 
\section{Conclusion and Limitations} \label{sec:conclusion}
% talk about safety

In this paper, we present a multimodal LLM, a.k.a. Chemma-RC, for chemical reaction condition prediction. Trained with Q\&A instruction datasets along with text-augmented corpus, graph, and SMILES representation, Chemma-RC effectively answers questions about reaction conditions. 
Even though the integration of the graph modality increases computational cost while offering limited performance gains in the current task, we believe it holds promise for more complex reaction scenarios. We plan to further explore cross-modal alignments, where graph-based contributions are expected to play a more significant role.
In the future, ensuring the safety and feasibility of generated reaction conditions is critical, especially when deploying the model in autonomous synthesis platforms. Second, the trade-off between computational efficiency and predictive performance warrants further investigation, particularly for scaling the model to broader chemical domains or real-time applications.

% \begin{acks}
% \end{acks}

\bibliographystyle{ACM-Reference-Format}
\bibliography{sample-base}

\newpage

\appendix
\section*{Appendix}
\section{Data Description} \label{app: data description}
% 先介绍format 再介绍size
% The reaction condition prediction task asks models to predict reaction conditions such as reagents, and can be categorized into two subtasks: reaction condition classification and reaction condition generation. For the classification task, we utilize the USPTO-Condition dataset. For the generation task, we USPTO\_500MT\_Condition dataset. 
% 我们的任务分为两类，分别用到两个数据集
% We formulate the reaction condition prediction task as two subtasks: reaction condition classification and reaction condition generation. The classification task is constructed based on the USPTO-Condition dataset, while the generation task is based on the \\
% USPTO\_500MT\_Condition dataset.

We introduce two large benchmark datasets, USPTO-Condition and USPTO\_500MT\_Condition to evaluate model performance on recommending reaction conditions.

\begin{figure}[h]
  \centering
  \includegraphics[width=0.9\linewidth]{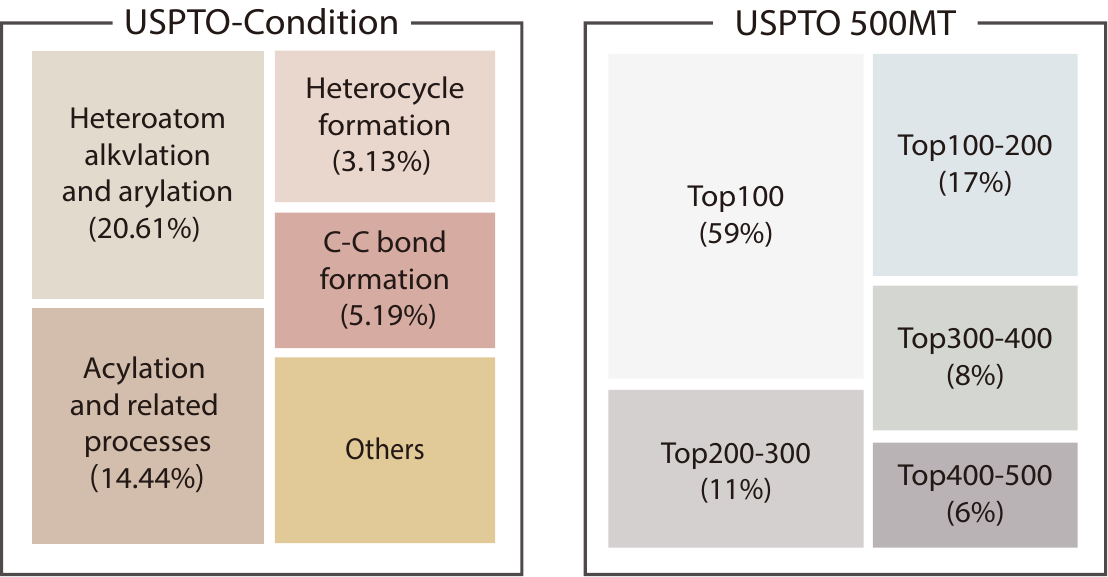}
  \caption{Reaction category composition of the USPTO-Condition dataset and the USPTO\_500MT\_Condition dataset.}
  \label{fig:dataset}
\end{figure}

% 谁提出来的，数据有哪些列，训练计划分的两种策略
The USPTO-Condition dataset, curated by Wang et al.~\cite{wang2023generic}, is commonly used in previous works~\cite{wang2023generic, qian2023predictive}. It comprises about 700,000 reaction data derived from the public USPTO data~\cite{lowe2012uspto}, with heteroatom alkylation and arylation reactions accounting for the majority, as shown in Figure~\ref{fig:dataset} (left). Each reaction entry consists of reaction reactants, products, and conditions in canonical SMILES format. As shown in Table~\ref{tab:data format}, the reaction conditions are categorized into five distinct types, including \textit{catalyst}, \textit{solvent 1}, \textit{solvent 2}, \textit{reagent 1}, and \textit{reagent 2}, and are presented in a consistent order, which facilitates a more structured prediction task. 

% data split strategy
We follow the data split strategy proposed by Wang~\cite{wang2023generic}, which randomly divides reactions into training, validation, and testing sets with a ratio of 8:1:1, and detailed statistics are presented in Table~\ref{tab:data format}. Furthermore, in section ~\ref{sec:data split}, for a fair comparison, we evaluate model performance under the time-based data split (TS) strategy proposed by Qian~\cite{qian2023predictive}, where the reactions collected before 2015 are categorized as the training set, reactions from 2015 as validation, and reactions from 2016 as testing. 

% 介绍uspto 500mt
The USPTO\_500MT\_Condition dataset, introduced by Lu et al.~\cite{lu2022unified}, collects about 110,000 reactions with the top-500 common reaction types sourced from the USPTO-MIT dataset~\cite{coley2017prediction}, and reactions with the top-100 reaction types constitute 59\% of the dataset, as illustrated in Figure~\ref{fig:dataset} (right). Each entry in the \\
USPTO\_500MT\_Condition dataset comprises reactants, products, and conditions. Notably, as shown in Table~\ref{tab:data format}, all reaction conditions in USPTO\_500MT\_Condition dataset are concatenated using dots and collectively labeled as \textit{reagents}, which results in an unstructured sequence generation task. 

We perform data cleaning on the USPTO\_500MT\_Condition dataset to ensure consistency and quality. Specifically, we canonicalize all molecular SMILES representations and remove reactions with more than six reagents. Following the data splitting strategy proposed by Lu~\cite{lu2022unified}, the dataset is randomly divided into training, validation, and test sets in an 8:1:1 ratio. Detailed statistics are reported in Table~\ref{tab:data format}.

%%%%%%%%%%%%%%%%%%%%%%%%%%%%%%%%%%%%%%%%%%%%%%%%%%%%%%%%%
\begin{table*}[!htbp]
\centering
\small
% \scriptsize
\caption{Data description of USPTO-Condition and USPTO\_500MT\_Condition datasets.}
% \renewcommand\arraystretch{1.2}
% \vspace{-2pt}
    % \resizebox{\linewidth}{!}{
% \setlength{\tabcolsep}{1mm}{
\begin{tabular}{cccccc}
\toprule
% \multicolumn{5}{c}{\textbf{USPTO-Condition}} \\
% \toprule
\multicolumn{1}{c}{Dataset} &  \multicolumn{1}{c}{Sample of conditions} & \multicolumn{1}{c}{Prediction type} & \multicolumn{1}{c}{Training} & \multicolumn{1}{c}{Validation} & \multicolumn{1}{c}{Testing}\\
\midrule
USPTO-Condition & [Zn],C1CCOC1,O,CO,[Cl-].[NH4+] & classification & 546,728 & 68,341 & 68,341\\
% \toprule
USPTO$\_$500MT$\_$Condition & CO.[Na+].CC(=O)O.[BH3-]CN & generation &  88,410 & 9,778 & 10,828\\
% \midrule
% \textbf{Parrot} & 13.8   &\\
% % \midrule
% % \rowcolor{mygray}
% \textbf{MM-RCR} & 25.9  & \\
\bottomrule
% \hline
\end{tabular}
% }
\label{tab:data format}
% \vspace{-10pt}
\end{table*}
%%%%%%%%%%%%%%%%%%%%%%%%%%%%%%%%%%%%%%%%%%%%%%%%%%%%%%%%%

%%%%%%%%%%%%%%%%%%%%%%%%%%%%%%%%%%%%%%%%%%%%%%%%%%%%%%%%

%%%%%%%%%%%%%%%%%%%%%%%%%%%%%%%%%%%%%%%%%%%%%%%%%%%%%%%%

% 讲对数据分布的分析
% 具体介绍数据集，有哪些特点。
To obtain a comprehensive understanding of data distribution, we perform an in-depth data analysis on the USPTO-Condition and USPTO\_500MT\_Condition datasets. 

% 第一个，分析了数据的非空比例
First, we calculate the non-empty count and non-empty proportion of each condition type. On the USPTO-Condition dataset, \textit{catalyst}, \textit{reagent 2} and \textit{solvent 2} condition types exhibit a high extent of sparsity, with non-empty entries occurring in less than 30\% of reactions, as shown in Table~\ref{tab:sparsity-USPTO-Condition}. It indicates that some reactions do not require multiple reagents and solvents, and the corresponding condition labels are therefore assigned as \textit{None}. On the USPTO\_500MT\_Condition dataset, since reaction conditions are represented as a single dot-concatenated string, all reactions are associated with non-empty condition labels.

% To obtain a comprehensive understanding of data distribution, we calculate the non-empty count and density of conditions and perform sparsity analyses on both USPTO-Condition and USPTO\_500MT\_Condition datasets. From the Table~\ref{tab:sparsity-USPTO-Condition}, we can infer that in the USPTO-Condition dataset, condition components including `Catalyst', `Solvent 2', and `Reagent 2' exhibit a high extent of sparsity, with non-empty density of fewer than 30\%. In contrast, there are more than 500,000 reaction entries with non-empty `Solvent 1' or `Reagent 1' compounds, underscoring their significance in chemical experimentation. For the USPTO\_500MT\_Condition dataset, all reaction entries have their corresponding non-empty condition molecules in the `Reagents' column.

%%%%%%%%%%%%%%%%%%%%%%%%%%%%%%%%%%%%%%%%%%%%%%%%%%%%%%%%

\begin{table}[htbp]
\centering
\caption{Sparsity analysis of the USPTO-Condition dataset.}
\scriptsize
\resizebox{\linewidth}{!}{
\begin{tabular}{cccccc}
\toprule
% \multicolumn{5}{c}{\textbf{USPTO-Condition}} \\
% \toprule
Non-empty & \multicolumn{1}{c}{catalyst} & \multicolumn{1}{c}{solvent 1} & \multicolumn{1}{c}{solvent 2} & \multicolumn{1}{c}{reagent 1} & \multicolumn{1}{c}{reagent 2} \\
\midrule
Count & 89,756 & 673,634 & 130,326 & 504,169 & 170,752 \\
% \toprule
Density & 13\% & 99\% & 19\% & 74\% & 25\% \\
\bottomrule
\end{tabular}
}
\label{tab:sparsity-USPTO-Condition}
\end{table}

%%%%%%%%%%%%%%%%%%%%%%%%%%%%%%%%%%%%%%%%%%%%%%%%%%%%%%%%

\begin{figure*}[!]
  \centering
  \includegraphics[width=0.7\linewidth]{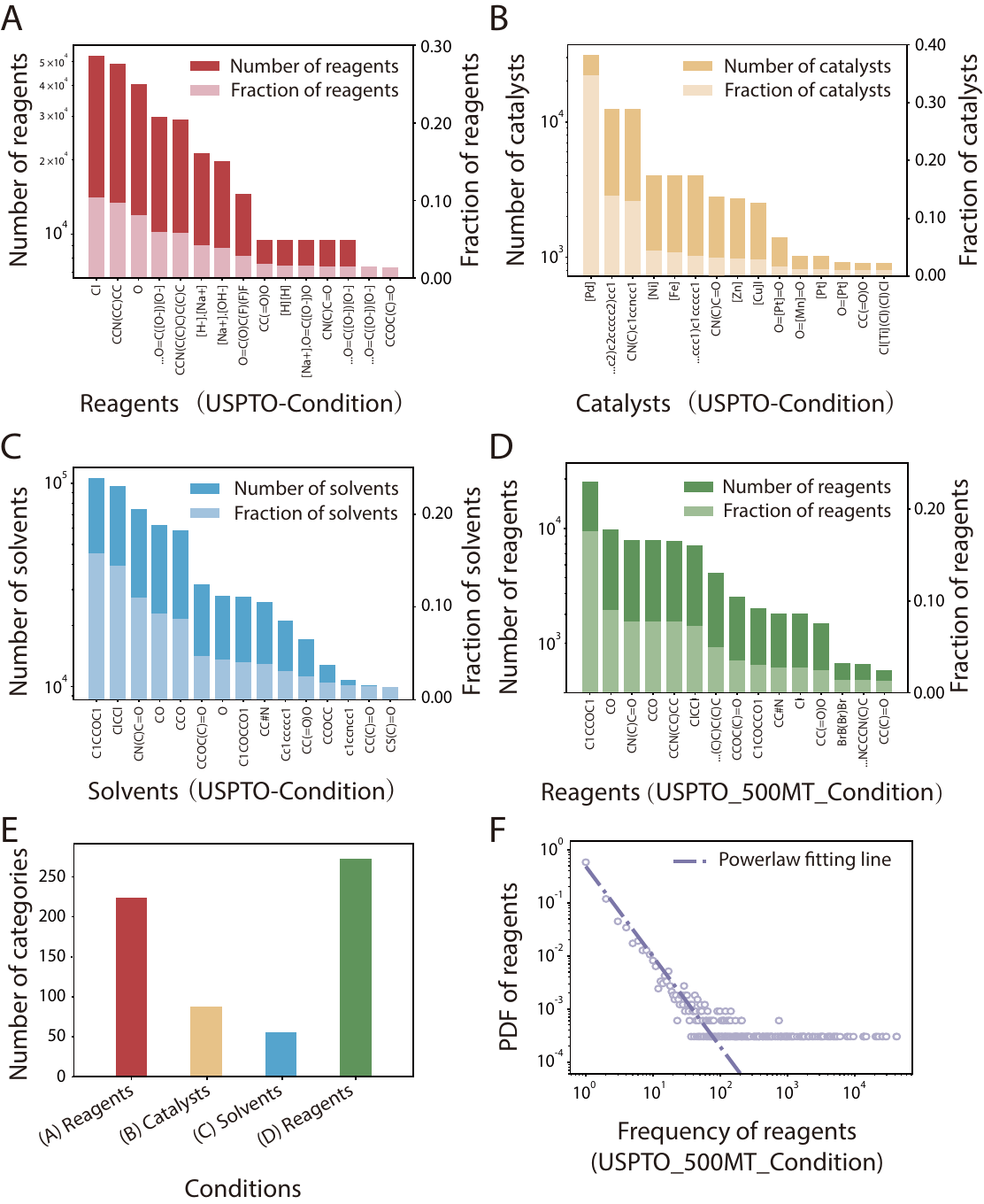}
  \caption{Distribution of types of reactions in the USPTO-Condition and USPTO\_500MT\_Condition datasets. \textbf{(A-D)} The bar charts of fifteen most common conditions in the two datasets, respectively. The deep color presents the decimal-scale proportion and the shallow color presents the log-scale count. \textbf{(E)} The number of distinct conditions in the two datasets, while (A)-(C) correspond to the USPTO-Condition dataset, and (D) corresponds to the USPTO\_500MT\_Condition dataset. \textbf{(F)} Power law fitting of the reagent distribution in the USPTO\_500MT\_Condition dataset, where the shallow points show the probability density and the deep dashed line shows the ideal power-law fitting. }
  % \caption{Reaction condition distribution in the USPTO-Condition and USPTO\_500MT\_Conditiond datasets. \textbf{(A-D)} The occurrence frequency of condition molecules across different datasets. \textbf{(E)} Counts of condition categories corresponding to the compounds. \textbf{(F)} Power-law fitting of the reagent distribution in the USPTO\_500MT\_Condition.}
  \label{fig:dataset-intro}
\end{figure*}

% 第二个，分析了数据的常见条件有哪些
Second, we explore the inner distribution characteristics across the two dataset, as illustrated in Figure~\ref{fig:dataset-intro}. Reaction conditions exhibit a high degree of diversity and imbalance in both datasets. In Figure~\ref{fig:dataset-intro}(F), we confirm that the occurrence frequency of reagents in the datasets follow a power-law distribution. The power-law distributions demonstrate the long-tail characteristics and a small number of categories account for the majority of the whole dataset. Such phenomenon is in light with the distribution of words in natural language, indicating that there is potential for tackling chemical tasks with natural language models.

\begin{table}[!h] 
\scriptsize
\centering
    \caption{Question templates generated by GPT-4.}
    %\vspace{5mm}
    \renewcommand\arraystretch{1.5}
    \setlength{\tabcolsep}{2.5
    mm}{
    \resizebox{\linewidth}{!}{
    \begin{tabular}{cc}
\toprule
   Task & Description\\
    \toprule
        Solvent prediction & \makecell{Could you suggest potential solvents that could have been used \\ in the given chemical reaction, taking into consideration their polarity \\ and compatibility with the reactants?} \\
    \midrule
        Reagent prediction & \makecell{Please suggest some possible reagents that could have been used \\ in the following chemical reaction.} \\
    \midrule
        Catalyst prediction & \makecell{Considering the chemical reaction in question, \\ which catalysts could be effective?  } \\
    \midrule
        Condition prediction (all) & \makecell{Given the current chemical reaction, what would be the appropriate \\ conditions to consider?} \\
    \bottomrule
    \end{tabular}}
    }
    \label{tab:question prompt}
\end{table}

% %%%%%%%%%%%%%%%%%%%%%%%%%%%%%%%%%%%%%%%%%%%%%%%%%%%%%%%
% \begin{table}[h]
% \small
% \centering
%     \caption{Data volume of USPTO-Condition and USPTO\_500MT\_Condition datasets.}
%     %\vspace{5mm}
%     \renewcommand\arraystretch{1.15}
%     \setlength{\tabcolsep}{4
%     mm}{
%     \begin{tabular}{ccccc}
% \toprule
%    Dataset & Training set & Validation set & Testing set\\
%     \toprule
%         USPTO-Condition & 546,728 & 68,341 & 68,341\\
%         %Atom Fingerprint + Context & 73.1 & 79.2 & 88.3 & 94.7\\
%         USPTO\_500MT\_Condition & 88,410 & 9,778 & 10,828\\
%         % USPTO\_500MT\_Condition & 89306 & 9881 & 10961\\

%     \bottomrule
%     \end{tabular}}
%     \label{tab:data-info}
% \end{table}
% 

%%%%%%%%%%%%%%%%%%%%%%%%%%%%%%%%%%%%%%%%%%%%%%%

% \begin{table}[htbp]
% \centering
% \small
% % \resizebox{\linewidth}{!}{
% \begin{tabular}{ccccccc}
% \toprule
% % \multicolumn{6}{c}{\textbf{USPTO\_500MT\_Condition}} \\
% % \toprule
% \textbf{USPTO\_500MT\_Condition} & \multicolumn{1}{c}{Reagent 1} & \multicolumn{1}{c}{Reagent 2} & \multicolumn{1}{c}{Reagent 3} & \multicolumn{1}{c}{Reagent 4} & \multicolumn{1}{c}{Reagent 5} & Reagent 6 \\
% \midrule
% \textbf{Non-empty count} & 108,913 & 92,749 & 51,783 & 19,191 & 5,710 & 1,412 \\
% % \toprule
% \textbf{Non-empty density} & 99\% & 85\% & 48\% & 18\% & 5\% & 1\% \\
% \bottomrule
% \end{tabular}
% \caption{USPTO\_500MT\_Condition}
% \label{tab:ds-USPTO_MT500_Condition}
% \end{table}

%%%%%%%%%%%%%%%%%%%%%%%%%%%%%%%%%%%%%%%%%%%%%%%%%%%%%%%%%%%%%%

\section{Training Settings} \label{app: training settings}
% This section offers a comprehensive summary of the dataset details, and training settings. In our study, the USPTO-Condition and USPTO\_500MT\_Condition datasets are split according to the ratio of 8:1:1, with the specific counts presented in Table~\ref{tab:data format}. 
% We leverage GPT-4 to generate distinct question templates for question prompts, following the policy of diversity, consistency, and completeness. The correctness of all question templates generated by GPT-4 is verified through manual review. The examples of the question templates are shown in Table~\ref{tab:question map}.

Within the model framework, Chemma-RC takes the 32-layer LLaMA-2-7b as the LLM backbone. For the reaction representation, we introduce Parrot~\cite{wang2023generic} to encode the reaction SMILES. For the graph-based reaction representation, we leverage R-GCN~\cite{schlichtkrull2018modeling}. For the text-based reaction representation, we retrieve the corresponding similar corpus and utilize LLaMA-2 as a text decoder. During the training process, the weight parameters of the graph and reaction encoders are frozen, while the modality alignment and part layers of LLaMA-2 are trainable. We utilize parameter-efficient LoRA for SFT, and the trainable parameters constitute approximately 0.3 billion out of the total 7 billion parameters. The multimodal SFT process is conducted with a batch size of 16 for fewer than 6 epochs over 48 hours, utilizing a GPU configuration of 2 $\times$ 48 GB NVIDIA A6000 GPUs. Inference is performed on a single 80 GB NVIDIA A800 GPU. 

% Besides, we utilize a pre-trained SMILES-to-text retriever proposed by Qian et al.~\cite{qian2023predictive} and extract the most similar unpaired corpus as the reaction text. Meanwhile, we introduce Parrot, a Bert-like model to encode the reaction SMILES. We leverage R-GCN~\cite{schlichtkrull2018modeling} to encode the molecules in the reaction, and the combination of reactant and product embeddings is considered as the reaction representation. In the training process, the encoders in all modalities are frozen. After the alignment of the representation space, the SMILES and the graph-based tokens have a length of 128 and 3, respectively. Additionally, the model employs the OneCycleLR as the learning rate scheduler, initializing the learning rate as 3e-5. The batch size is set to 16, with less than 6 epochs 48 hours in training. The GPU configuration is 8 $\times$ 80G A800.

\section{Details of Modality Alignment} \label{app: alignment}
% Meanwhile, we also take the instruction prompts as the input, and generate the text tokens ${\mathbf Q}_t\in\mathbb{R}^{M\times C}$, where $M$ denotes the number of queries. As depicted in Figure~\ref{algo:code}, In this way, the text tokens ${\mathbf Q}_t\in\mathbb{R}^{M\times C}$ decoded by LLMs contains highlighted reaction cues that are most related to the user instruction.
%
For the reaction condition prediction task, the representation of the reaction is extracted by encoders, and the text representation is tokenized by LLMs. However, fusing two types of representation introduces inductive biases issues~\cite{baltruvsaitis2018multimodal, jaegle2021perceiver}. To effectively fuse representations from multiple modalities, we propose the use of a projection module, the Perceiver \cite{jaegle2021perceiver}, for modality alignment (Figure~\ref{fig:framework}). This module employs latent queries to align graph and SMILES tokens with text-related tokens, such as question prompts and a text-augmented corpus. We show the pseudo-code for modality projection in Algorithm.~\ref{algo:code}.

\begin{algorithm}[h]
\caption{Pseudo code for modality projection.}
\label{algo:code}
% \algcomment{\fontsize{9pt}{0em}\selectfont \texttt{word\_proj}, \texttt{perceiver\_proj}: predefined linear and transformer-based projectors, respectively.
% }

\definecolor{codeblue}{rgb}{0.25,0.5,0.5}
\lstset{
  backgroundcolor=\color{white},
  basicstyle=\fontsize{9pt}{9pt}\ttfamily\selectfont,
  columns=fullflexible,
  breaklines=true,
  captionpos=b,
  commentstyle=\fontsize{9pt}{9pt}\color{codeblue},
  keywordstyle=\fontsize{9pt}{9pt},
%  frame=tb,
}
\begin{lstlisting}[language=python]
# B: batch size; C: channel size; n: content shape
# M: query length; N: shape of flatten reaction tokens;
# text_q: text query in shape (B, M, C)
# react_embed: reaction embedding in shape (B, N, C)
# word_embed: word embedding in shape (B, vocab_size, C)

# Key part 1: map transformer-based reaction feature 
word_embed = self.word_proj(word_embed)
word_embed = word_embed.repeat(react_embed.size()[0], 1, 1)
react_embed = torch.cat([react_embed, word_embed], dim=1)
smiles_react_tokens = linear_layer(react_embed) # to make 128 tokens

# Key part 2: map graph-based reaction features 
graph_embed = self.word_proj(graph_embed)
graph_react_tokens = linear_layer(graph_embed) # to make 3 tokens

# Key part 3: 
reaction_tokens = torch.cat([smiles_react_tokens, graph_react_tokens], dim=1)

# Key part 4: modality projection
reaction_tokens_from_smiles = self.perceiver_proj_smiles(smiles_react_tokens)
reaction_tokens_from_graphs = self.perceiver_proj_graphs(graph_react_tokens)

# concat token
final_token = torch.cat([reaction_tokens_from_smiles, reaction_tokens_from_graphs, text_q], dim=1)

\end{lstlisting}
\end{algorithm}

\section{Model Performance} \label{app: model performance}
A chemical reaction can be represented as the transformation of a sequence of characters (reactants, conditions) into another sequence (products), with compounds connected by special characters, such as `$\textgreater\textgreater$'. This structure makes sequence-to-sequence models, such as the Transformer, well-suited for predictive modeling of reaction representation \cite{schwaller2019molecular, irwin2022chemformer}. However, existing SMILES-based Transformer models for reaction representation encounter limitations in various aspects, particularly with respect to atom permutations and the interpretability of reaction mechanisms.
% why LLM can work 可以学到机理，无需数据增强
% Towards these points, LLMs can be a potential solution owing to its advances: (i) ingested with large corpus about chemistry, LLMs acquire chemical knowledge akin to the learning process of chemists~\cite{openai2023gpt}; (ii) pre-trained with extensive reaction data, LLMs might be empowered the capability of understanding the mechanism of reactions, independent of DFT calculation~\cite{ai4science2023impact}.
Consequently, our proposed Chemma-RC fuses data from diverse sources including corpus, SMILES and graphs of molecules to present a comprehensive view of the reaction.
% 介绍数据集，介绍baseline方法，结果分析
We assess the performance of our proposed Chemma-RC and the aforementioned baseline methods for reaction condition prediction. The top-$N$ reaction condition prediction accuracy on USPTO-Condition and USPTO\_500MT\_Condition datasets are presented in Table~\ref{tab:performance-condition} and Table~\ref{tab:performance-500mt}, respectively. We introduce several comparative baseline methods.
\begin{enumerate}

% 关注下哪些是我们抄的表，哪些是复现的吧

\item rxnfp LSTM~\cite{gao2018using}. This method introduces a reaction representation based on Morgan fingerprints, defined as the difference between the fingerprint vectors of the products and the reactants. We report reproduced results from Wang et al.~\cite{wang2023generic} in Table~\ref{tab:performance-condition}, and results from Qian et al.~\cite{qian2023predictive} in Table~\ref{tab: data split}.  

\item Parrot~\cite{wang2023generic}. This method leverages a powerful attention-based model architecture to encode the reaction. We report evaluation results from Wang et al.~\cite{wang2023generic} in Table~\ref{tab:performance-condition}. Further, we follow the training setting in the paper, and test Parrot's performance on the USPTO\_500MT\_Condition dataset in Table~\ref{tab:performance-500mt}. 

\item TextReact~\cite{qian2023predictive}. This method introduces relevant corpus retrieved from literature to enhance the molecular representation of the reaction based on SMILES. For a fair comparison, we exclude the gold texts paired with each chemical input during both training and evaluation. Our reproduced results are reported in Table~\ref{tab:performance-condition}, where we referred as TextReact\_s. 

\item DeepSeek-V2~\cite{liu2024deepseek}, GPT-4o~\cite{achiam2023gpt}, LLaMA3-70B~\cite{llama-2}. They are general-purpose generative large language models pretrained on massive corpora of diverse text data, which sparked significant interest in the field of AI for chemistry. In Table~\ref{tab:performance-500mt}, we utilize Ollama and OpenAI API to evaluate model performance under three settings: zero-shot, one-shot and five-shot. 

\item ChemDFM~\cite{zhao2025developing}. It is a pioneering LLM for chemistry trained on 34B tokens from chemical literature and textbooks, and fine-tuned using 2.7M instructions. We download the open-sourced ChemDFM weights and evaluate model performance under zero-shot, one-shot and five-shot settings. We report our evaluation result in Table~\ref{tab:performance-500mt}. 

\item Reagent Transformer~\cite{andronov2023reagent}. This method leverages Molecular Transformer~\cite{schwaller2019molecular} to tackle the task of reagent prediction. We reproduce the model with the training settings reported in the paper and evaluate model performance on the USPTO\_500MT\_Condition dataset in Table~\ref{tab:performance-500mt}.

\item Reaction GCNN~\cite{maser2021multilabel}. This method proposes a machine-learned ranking model to predict the condition set. We reproduce the model with the training settings reported in the paper and evaluate model performance in the \\
USPTO\_500MT\_Condition dataset on in Table~\ref{tab:performance-500mt}.

\item nach0~\cite{livne2024nach0}. This method is a multi-domain and multi-task encoder-decoder LLM pre-trained on unlabeled text from scientific literature, patents, and molecule strings to incorporate a range of chemical and linguistic knowledge. In Table~\ref{tab:performance-500mt}, we report evaluation  results from the paper. 

\item TextReact~\cite{qian2023predictive} variants: rxnfp retrieval, Transformer, ChemBERTa, TextReact(gr). rxnfp retrieval takes the conditions of the most similar reactions in the training set as the prediction. Transformer uses the same architecture as the TextReact predictor. ChemBERTa is same as the Transformer baseline except that the encoder is pretrained on external SMILES data. TextReact(gr) removes the gold corpus in the evaluation process. In Table~\ref{tab: data split}, we report model performance from Qian et al.~\cite{qian2023predictive}.

% \item rxnfp retrieval. It takes the conditions of the most similar reactions in the training set as the prediction. Similar reactions are determined based on the $L_2$
% distance of reaction fingerprints.

% \item Reaction GCNN~\cite{maser2021multilabel}. This method proposes a machine-learned ranking model to predict the set of conditions used in a reaction as a binary vector.

% \item Parrot~\cite{wang2023generic}. This method leverages the attention-based model architecture to encode the reaction and design a training methodology specifically to augment the reaction center.

% \item TextReact~\cite{qian2023predictive}. It aims to enhance reaction representation by introducing relevant corpus retrieved from literature into sequence-to-sequence Transformers.

% \item Transformer. It uses the same architecture as the TextReact predictor. This baseline represents the state-of-the-art model that only takes chemistry input.

% \item ChemBERTa~\cite{chithrananda2020chemberta}. It is same as the Transformer baseline except that the encoder is pre-trained on external SMILES data.

% \item Reagent Transformer~\cite{andronov2023reagent}. This method leverages Molecular Transformer,~\cite{schwaller2019molecular} a state-of-the-art model to tackle the task of reagent prediction.

% \item nach0~\cite{livne2024nach0}. This method is a multi-domain and multi-task encoder-decoder LLM pre-trained on unlabeled text from scientific literature, patents, and molecule strings to incorporate a range of chemical and linguistic knowledge. 

\end{enumerate}

% To have a comprehensive overview of the recommendation performance, we visualize the prediction results of USPTO-Condition and USPTO\_500MT\_Condition datasets, as described in Table~\ref{tab:performance-condition}, ~\ref{tab:performance-500mt}. Specifically, we draw radar charts of our model and other competitive models, which are presented in Figure~\ref{fig:pf-radar}. For the USPTO-Condition dataset, we reproduce Parrot, RCR, and TextReact. Then, we plot the top-3 predicting accuracy of different conditions (catalyst, solvent 1, solvent 2, reagent 1, and reagent 2), as depicted in Figure~\ref{fig:pf-radar} (left). For the USPTO\_500MT\_Condition dataset, we recommend reagents in SMILES sequence and take Parrot, Reagent Transformer, and Reaction GCNN as comparative methods. For more intuition, we visualize top-1, 3, 5, and 10 exactly matched accuracy in log scale, which is shown in Figure~\ref{fig:pf-radar} (right). From the charts, we can see that our model covers the largest area of the performance circle in both datasets, indicating that Chemma-RC markedly outperforms other competitive models.

\subsection{Generalization Performance} \label{app: general}
% employ the test set of USPTO\_500MT\_Condition for validation of
In order to validate the out-of-domain performance of Chemma-RC, we employ Chemma-RC trained on the USPTO\_500MT\_Condition to test on the USPTO-Condition. The evaluation strategy includes three specific training conditions: reagents, catalysts, and solvents. We adopt a metric of \textbf{partial matched accuracy} to illustrate the generalization capability of Chemma-RC. Different from the complete matched accuracy that requires perfect matching between predictions and labels, the partial matched accuracy is more suitable to test the generalization capacity, which focuses more on whether the predicted results match a substitutable part of the ground truth. For example, if the predicted result is `[Na+].[OH-]' and the condition label is `CO.[Na+].[OH-]', we consider that the prediction partially matches the ground truth, but not completely. The evaluation strategy includes three specific training conditions: reagents, catalysts, and solvents. Table~\ref{tab:cross-data} reports the top-1 partial match accuracy for each condition prediction. From the results we can see that, Chemma-RC achieves a top-1 partial matched accuracy of 67.1\% and 58.1\%, respectively. This relatively high accuracy indicates that solvents and reagents have more consistent characteristics that the model can learn effectively from USPTO\_500MT\_Condition and apply to USPTO-Condition. In contrast, The model's performance in predicting catalysts demonstrates a lower top-1 partial match accuracy at 89.9\%. 

\begin{table}[h]
% \small
\scriptsize
\centering
    \caption{The top-1 partial matched accuracy of Chemma-RC under OOD setting.}
    %\vspace{5mm}
    \renewcommand\arraystretch{1.15}
    \setlength{\tabcolsep}{4
    mm}{
    \resizebox{\linewidth}{!}{
    \begin{tabular}{cc}
\toprule
   Evaluation strategy (train $\to$ test) & Accuracy (\%) \\
    \midrule
    USPTO\_500MT\_Condition $\to$ USPTO-Condition (reagent) & 67.1 \\
    % \toprule
    USPTO\_500MT\_Condition $\to$ USPTO-Condition (catalyst) & 89.9  \\
    USPTO\_500MT\_Condition  $\to$ USPTO-Condition (solvent) &  58.1 \\
    \bottomrule
    \end{tabular}}
    }
    \label{tab:cross-data}
\end{table}

%%%%%%%%%%%%%%%%%%%%%%%%%%%%%%%%%%%%%%%%%%%%%%%%%%%%%%%%%
% \begin{table}{htbp}
%     \centering
%     \captionof{table}{
%     \label{tab:cross-data}}
%     % \renewcommand\arraystretch{1.2}
%     \vspace{-5pt}
%     \resizebox{\linewidth}{!}{
%     \begin{tabular}{cc}
%     \toprule
%     % \multicolumn{5}{c}{\textbf{USPTO-Condition}} \\
%     % \toprule
%     Evaluation strategy (train $\to$ test) & Acc (\%) \\
%     \midrule
%     USPTO\_500MT\_Condition $\to$ USPTO-Condition (reagent) & 67.1 \\
%     % \toprule
%     USPTO\_500MT\_Condition $\to$ USPTO-Condition (catalyst) & 89.9  \\
%     USPTO\_500MT\_Condition  $\to$ USPTO-Condition (solvent) &  58.1 \\
%     % \midrule
%     % \textbf{Parrot} & 13.8   &\\
%     % % \midrule
%     % % \rowcolor{mygray}
%     % \textbf{Chemma-RC} & 25.9  & \\
%     \bottomrule
%     % \hline
%     \end{tabular}}
%     \vspace{-5pt}
% \end{table}

%%%%%%%%%%%%%%%%%%%%%%%%%%%%%%%%%%%%%%%%%%%%%%%%%%%%%%%%%

% Further, we visualize the predicted results which can be seen in Figure~\ref{fig:rcr-case-study}. We select two reaction cases for analysis. In case 1, Toluene was not predicted by Chemma-RC. In case 2, 1,4-Dioxane and 1-(diphenylphosphaneyl)cyclopenta-2,4-dien-1-ide were predicted. However, it is confirmed that Toluene and 1,4-Dioxane are common solvents, and 1-(diphenylphosphaneyl)cyclopenta-2,4-dien-1-ide is frequently used as a ligand. Therefore, we do not categorize these as failed cases because the model successfully predicts all the reagents in the labels and avoids predicting other conditions.

Chemma-RC can successfully distinguish reagents from the combination of all conditions in a reaction. Additionally, training Chemma-RC on USPTO-Condition, a larger chemical reaction dataset, further enhances its ability to akin chemical knowledge.

\subsection{Case Study}
\begin{figure*}[htbp]
  \centering
  \includegraphics[width=\linewidth]{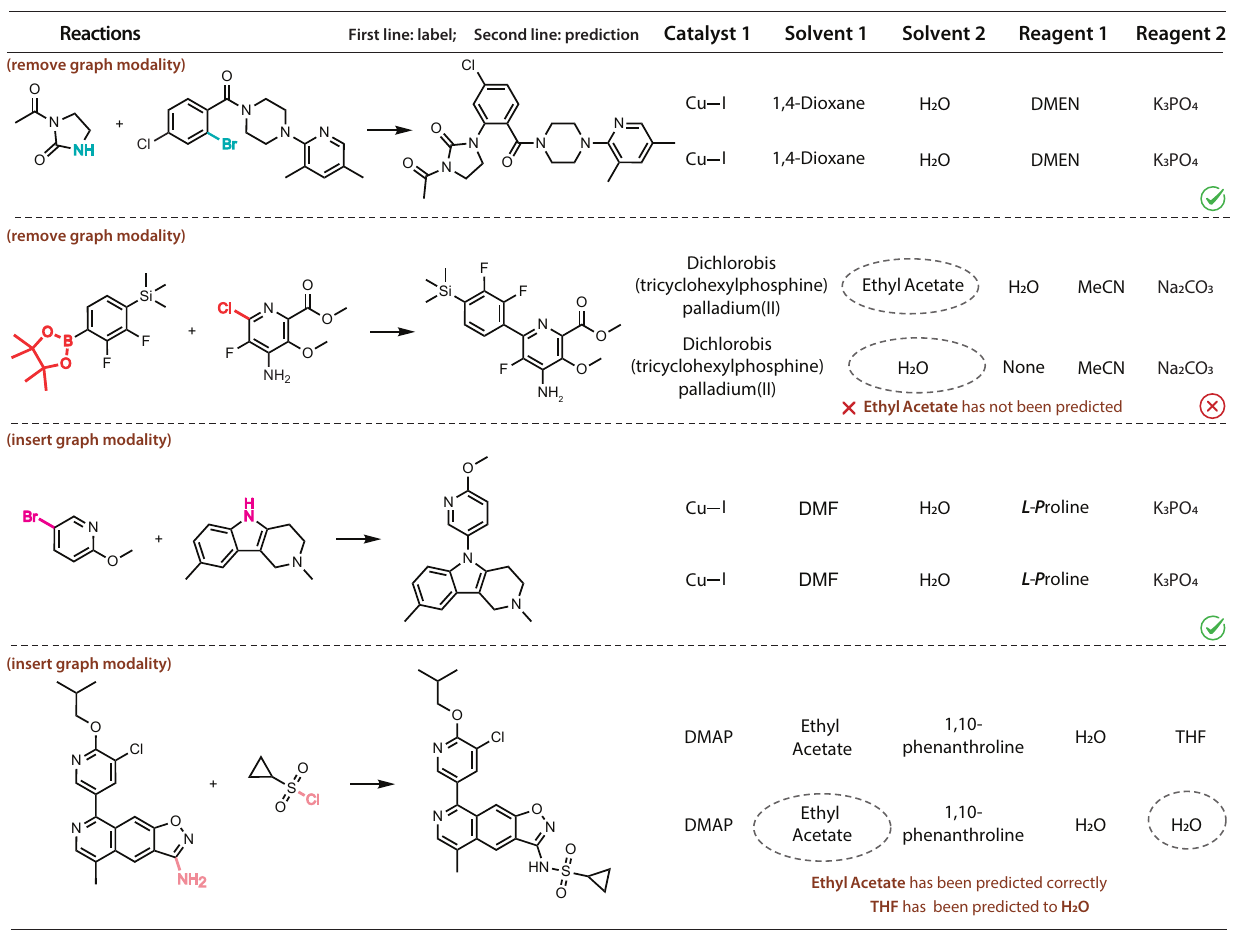}
  \caption{Visualization of generated conditions on four reactions. We select four Suzuki–Miyaura cross-coupling reactions to present the performance of condition prediction. The reaction centers and leaving groups are highlighted in different colors. }
  \label{fig:case-study-baseline}
\end{figure*}

In this section, we select four cross-coupling reactions from USPO-Condition datasets for performance validation. We visualize the predicted results in Figure~\ref{fig:case-study-baseline}. As depicted in Figure~\ref{fig:case-study-baseline}, the reaction centers and leaving groups are highlighted in different colors. For C--N cross-coupling reactions (the first and the third row), Chemma-RC can predict all conditions precisely. For C--C bond formation and Formylation reactions (the second and the fourth row), Chemma-RC fails to predict Ethyl Acetate (the second case) and THF (the fourth case). The reason why Chemma-RC is less effective for these reactions is that the data volume of C--C bond formation reactions in the USPTO-Condition dataset is only 5\%, as shown in Figure~\ref{fig:dataset}. This limited representation constrains the model's ability to learn the patterns associated with C--C bond formation reactions. Consequently, Chemma-RC lacks sufficient training examples to capture and generalize the underlying reaction mechanisms accurately. The scarcity of diverse and representative data hampers its effectiveness, leading to a lower precision in predicting these types of reactions.

%%%%%%%%%%%%%%%%%%%%%%%%%%%%%%%%%%%%%%%%%%%%%%%%%%%%%%%%%
\begin{figure*}[htbp]
  \centering
  \includegraphics[width=\linewidth]{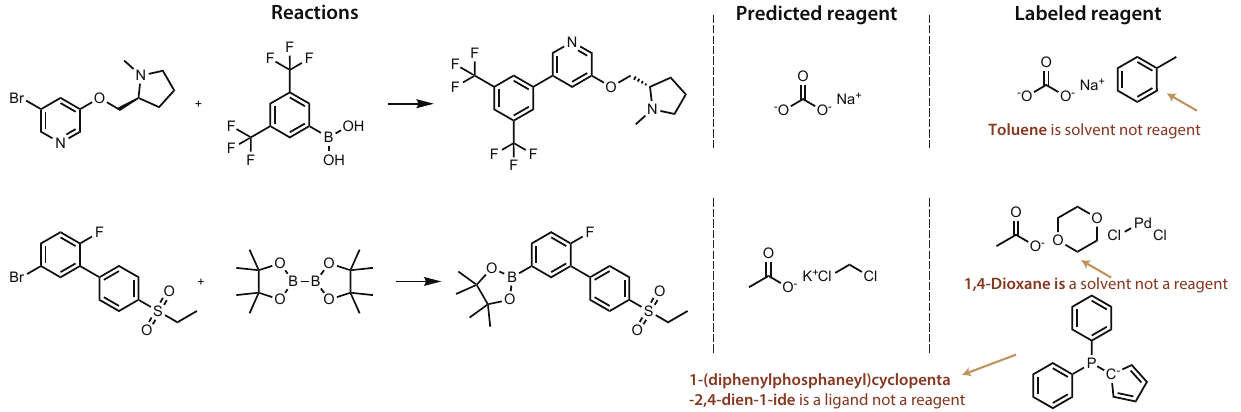}
  \caption{Visualization of recommended conditions on two reactions. In case 1, Toluene was not predicted by Chemma-RC. In case 2, 1,4-Dioxane and 1-(diphenylphosphaneyl)cyclopenta-2,4-dien-1-ide were predicted. However, it is confirmed that Toluene and 1,4-Dioxane are common solvents, and 1-(diphenylphosphaneyl)cyclopenta-2,4-dien-1-ide is frequently used as a ligand. Therefore, we do not categorize these as failed cases because the model successfully predicts all the reagents in the labels and avoids predicting other conditions.}
  \label{fig:rcr-case-study}
  % \vspace{-10pt}
\end{figure*}

%%%%%%%%%%%%%%%%%%%%%%%%%%%%%%%%%%%%%%%%%%%%%%%%%%%%%%%%%
Further, we visualize the predicted results on OOD datasets in Figure~\ref{fig:rcr-case-study}. We select two reaction cases for analysis. In case 1, Toluene is not predicted by Chemma-RC. In case 2, 1,4-Dioxane and 1-(diphenylphosphaneyl)cyclopenta-2,4-dien-1-ide are predicted. However, it is confirmed that Toluene and 1,4-Dioxane are common solvents, and 1-(diphenylphosphaneyl)cyclopenta-2,4-dien-1-ide is frequently used as a ligand. Therefore, we do not categorize these as failed cases because the model successfully predicts all the reagents in the labels and avoids predicting other conditions.

\section{Reproducibility Statement}

To ensure the reproducibility of our work, we have used datasets which have been published in \cite{wang2023generic, lu2022unified}, and the data links are as follows: \href{https://yzhang.hpc.nyu.edu/T5Chem/}{USPTO\_500MT\_Condition} and \href{https://drive.google.com/uc?id=1aX70qzZrJ9TZ9KpqnvUVR8WBxiTwXOsI}{USPTO-Condition}. Additionally, the code base for Chemma-RC is available as an anonymous repository for continuous development: \href{https://anonymous.4open.science/r/Chemma-RC-submission-5600/}{https://anonymous.4open \\.science/r/Chemma-RC-submission-5600/}.

%%%%%%%%%%%%%%%%%%%%%%%%%%%%%%%%%%%%%%%%%%%%%%%%%%%%%%%%%
%%%%%%%%%%%%%%%%%%%%%%%%%%%%%%%%%%%%%%%%%%%%%%%%%%%%%%%%%
\begin{figure*}[!t]
  % \vspace{-90pt}
  \centering
  \includegraphics[width=0.9\linewidth]{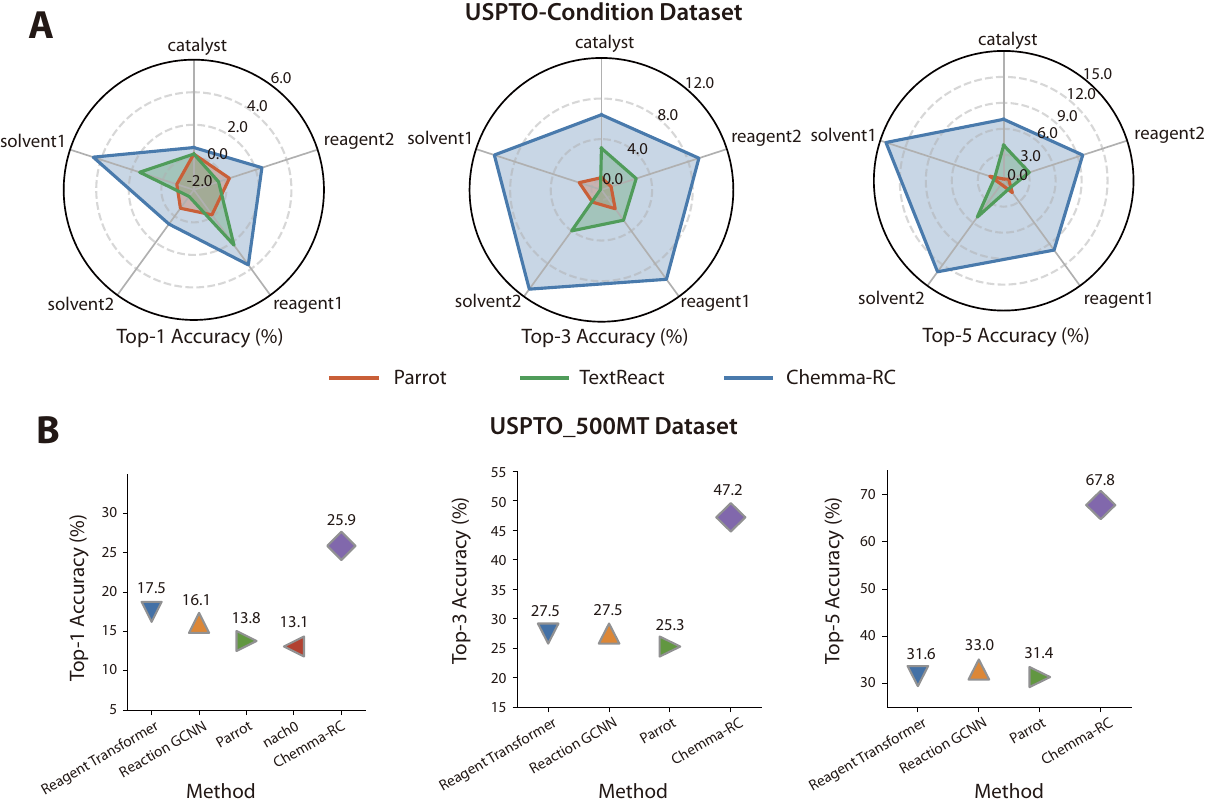}
  \caption{The visualization of relative performance enhancement. (A) relative performance enhancement on the USPTO-Condition dataset. (B) Performance evaluation for three baseline methods: Parrot (red), TextReact (green), and Chemma-RC (blue) on USPTO\_500MT\_Condition datasets.}
  \label{fig:pf-radar}
  % \vspace{-30pt}
\end{figure*}

%%%%%%%%%%%%%%%%%%%%%%%%%%%%%%%%%%%%%%%%%%%%%%%%%%%%%%%%%
\begin{figure*}[!t]
  % \vspace{-90pt}
  \centering
  \includegraphics[width=0.99\linewidth]{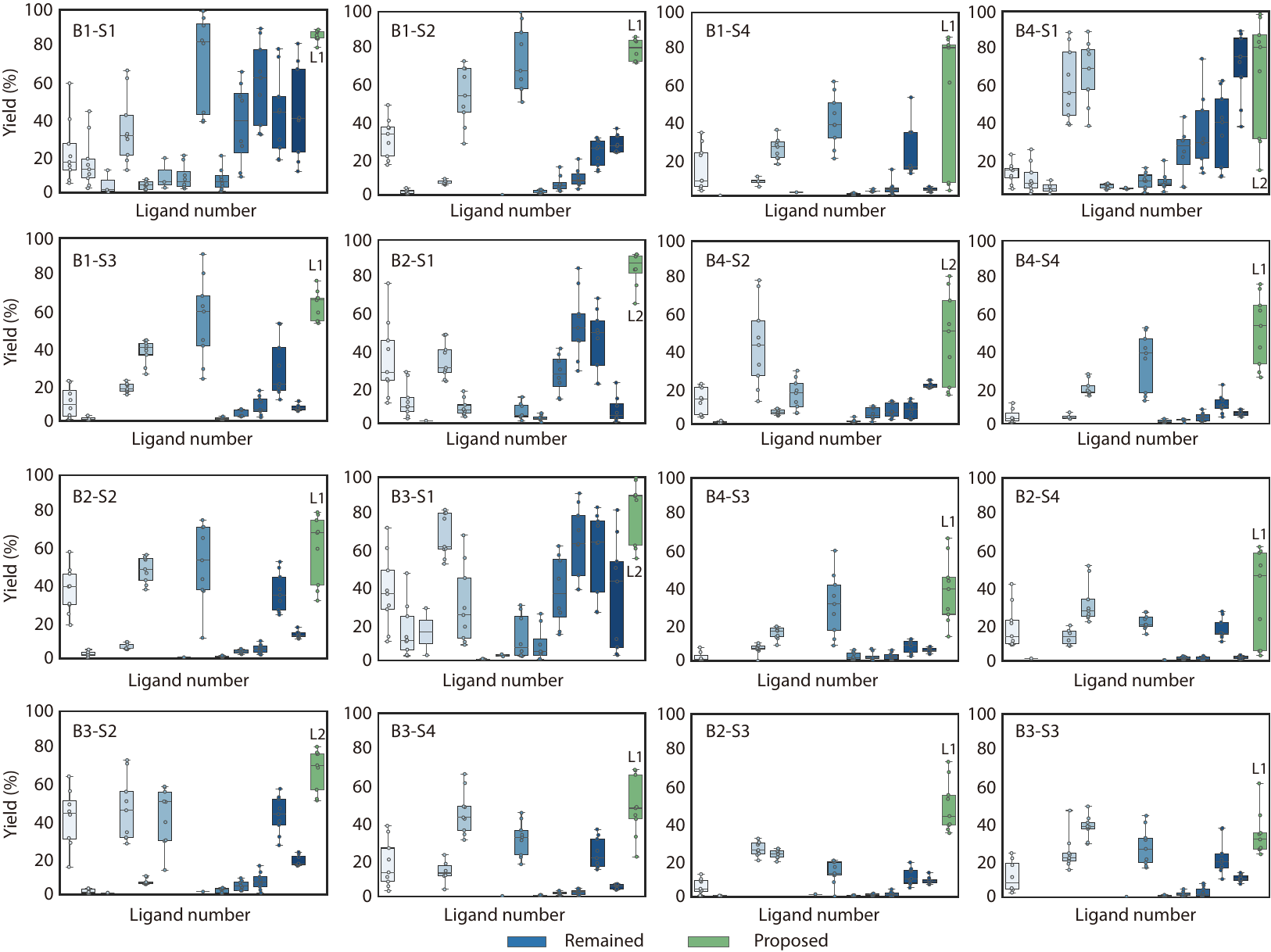}
  \caption{Performance evaluation of ligand generation.}
  \label{fig:ligand recommendation1}
  % \vspace{-30pt}
\end{figure*}

\begin{table*}[htb]
\centering
\scriptsize
\caption{Performance evaluation of Chemma-RC on specific reaction types on USPTO-Condition dataset. }
\resizebox{\linewidth}{!}{
\begin{tabular}{p{5.3cm}|l|c|c|c}
\toprule
\textbf{Reaction SMARTS} & \textbf{Reaction Name} & \textbf{Exact Match (\%)} & \textbf{Part Match (\%)} & \textbf{Recall (\%)} \\
\midrule
\rowcolor{cyan!10}
\multicolumn{5}{l}{\textbf{\textit{More complex reaction types}}} \\
\parbox[t]{6cm}{
[O;D1;H0:2]=[c;H0;D3;+0:1]1:[nH;D2;+0:3]:[c;H0;D3;+0:4]\\
(-[c;H0;D3;+0:5](:[c:6]):[c:7]):[n;H0;D2;+0:8]:[o;H0;D2;\\
+0:9]:1>>O-[C;H0;D3;+0:1](-[O-])=[O;D1;H0:2].[N;H0;D1;+0:\\
3][C;H0;D2;+0:4]-[c;H0;D3;+0:5](:[c:6]):[c:7].[NH3+;D1:\\
8]-[OH;D1;+0:9]} & Ring cleavage (retro 1,2,4-oxadiazole synthesis) & 66.13 & 100 & 88.46 \\

\parbox[t]{6cm}{
[C:3]-[O;H0;D2;+0:4]-[C;H0;D3;+0:1](=[O;D1;H0:2])-[NH;D2;\\
+0:5]-[c:6]>>Cl-C(-Cl)(-Cl)-O-[C;H0;D3;+0:1](=[O;D1;H0:2])-\\
O-C(-Cl)(-Cl)-Cl.[C:3]-[OH;D1;+0:4].[NH2;D1;+0:5]-[c:6]} & Carbamate cleavage (urethane deprotection) using phosgene or triphosgene & 100 & 100 & 100 \\
% \raggedright\arraybackslash\texttt{[C:3]-[O;H0;D2;+0:4]-[C;H0;D3;+0:1](=[O;D1;H0:2])-[NH;D2;+0:5]-[c:6]>>Cl-C(-Cl)(-Cl)-O-[C;H0;D3;+0:1](=[O;D1;H0:2])-O-C(-Cl)(-Cl)-Cl.[C:3]-[OH;D1;+0:4].[NH2;D1;+0:5]-[c:6]} & Carbamate cleavage (urethane deprotection) using phosgene or triphosgene & 100 & 100 & 100 \\
\parbox[t]{6cm}{[C:2]-[S;H0;D3;+0:3](=[O;H0;D1;+0:1])-[c:4]>>Cl-c1:c:c:c:c\\
(-C(=O)-O-[OH;D1;+0:1]):c:1.[C:2]-[S;H0;D2;+0:3]-[c:4]} & Reductive desulfonylation + ester formation (chlorinated benzoic acid) & 72.73 & 95.45 & 90.00 \\
\midrule
\rowcolor{cyan!10}
\multicolumn{5}{l}{\textbf{\textit{Less complex reaction types}}} \\
\parbox[t]{6cm}{[\#7;a:4]:[c:3]:[nH;D2;+0:1]:[c:2]>>C-c1:c:c:c(-S(=O)(=O)-[\\
n;H0;D3;+0:1](:[c:2]):[c:3]:[\#7;a:4]):c:c:1} & Sulfonylation of an N–H heterocycle & 61.54 & 92.31 & 86.49 \\
\parbox[t]{6cm}{[C:5]=[C:4]-[C:2](=[O;D1;H0:3])-[OH;D1;+0:1]>>C-[O;H0;D2;\\
+0:1]-[C:2](=[O;D1;H0:3])-[C:4]=[C:5]} & Fischer esterification of $\alpha,\beta$-unsaturated carboxylic acid & 40.00 & 96.00 & 81.00 \\
\parbox[t]{6cm}{[C:2]-[NH2;D1;+0:1]>>O=C1-[N;H0;D3;+0:1](-[C:2])-C(=O)-c2:\\
c:c:c:c:c:2-1} & Condensation of amine with phthalic anhydride to form phthalimide & 48.72 & 94.87 & 86.11 \\
\bottomrule
\end{tabular}
}
\label{tab:reaction-type-performance}
\end{table*}

%%%%%%%%%%%%%%%%%%%%%%%%%%%%%%%%%%%%%%%%%%%%%%%%%%%%%%%%%
\begin{table*}[htb]
\centering
% \scriptsize
% \renewcommand\arraystretch{1}
% \setlength{\tabcolsep}{1.8mm}
\caption{Performance comparison across different numbers of reaction conditions on USPTO\_500MT\_Condition dataset.}
\resizebox{0.8\linewidth}{!}
{
\begin{tabular}{c|c|ccccc}
\toprule
\textbf{Condition numbers} & \textbf{Frequency} & \textbf{Exact acc (\%)} & \textbf{Partial acc (\%)} & \textbf{Recall (\%)} & \textbf{Precision (\%)} \\
\midrule
1 & 1622 & 33.17 & 56.91 & 56.58 & 34.24 \\
2 & 4026 & 32.66 & 78.71 & 60.10 & 54.71 \\
3 & 3258 & 19.00 & 85.60 & 56.46 & 63.68 \\
4 & 1326 & 20.81 & 89.22 & 57.43 & 70.56 \\
5 & 440  & 20.91 & 91.59 & 57.83 & 73.20 \\
6 & 156  & 27.56 & 91.03 & 60.77 & 77.25 \\
\bottomrule
\end{tabular}
}
\label{tab:reagent-count-performance}
\end{table*}
%%%%%%%%%%%%%%%%%%%%%%%%%%%%%%%%%%%%%%%%%%%%%%%%%%%%%%%%%









\end{document}